\documentclass[sigconf]{acmart}

\AtBeginDocument{%
  \providecommand\BibTeX{{%
    \normalfont B\kern-0.5em{\scshape i\kern-0.25em b}\kern-0.8em\TeX}}}

\setcopyright{acmcopyright}
\copyrightyear{2018}
\acmYear{2018}
\acmDOI{XXXXXXX.XXXXXXX}

\acmConference[Conference acronym 'XX]{Make sure to enter the correct
  conference title from your rights confirmation emai}{June 03--05,
  2018}{Woodstock, NY}
\acmPrice{15.00}
\acmISBN{978-1-4503-XXXX-X/18/06}

\usepackage{xspace}

\newcommand{\model}{\textsc{CodeS}\xspace}
\newcommand{\smodel}{\textsc{CodeS}\xspace}

\NewDocumentCommand{\fanj}{mO{}}{\textcolor{blue}
{\textsuperscript{\textit{fanj}}\textsf{\textbf{\small[#1]}}}}

\newcommand{\stitle}[1]{\vspace{1.2pt}\noindent{\bf #1}}

\usepackage{multirow}
\usepackage{bbding}
\usepackage{pifont}
\usepackage{xcolor}
\usepackage{framed}
\usepackage[linesnumbered,ruled,vlined]{algorithm2e}
\definecolor{shadecolor}{rgb}{0.9,0.9,0.9}

\begin{document}

\title{\model: Towards Building Open-source Language Models for Text-to-SQL}

\author{Haoyang Li}
\affiliation{%
  \institution{Renmin University of China, China}
  \country{}
}
\email{lihaoyang.cs@ruc.edu.cn}

\author{Jing Zhang}
\affiliation{%
  \institution{Renmin University of China, China}
  \country{}
}
\email{zhang-jing@ruc.edu.cn}
\authornote{Jing Zhang is the corresponding author.}

\author{Hanbing Liu}
\affiliation{%
  \institution{Renmin University of China, China}
  \country{}
}
\email{liuhanbing@ruc.edu.cn}

\author{Ju Fan}
\affiliation{%
  \institution{Renmin University of China, China}
  \country{}
}
\email{fanj@ruc.edu.cn}

\author{Xiaokang Zhang}
\affiliation{%
  \institution{Renmin University of China, China}
  \country{}
}
\email{zhang2718@ruc.edu.cn}

\author{Jun Zhu}
\affiliation{%
  \institution{AI-Finance, China}
  \country{}
}
\email{zhujun@ai-finance.cn}

\author{Renjie Wei}
\affiliation{%
  \institution{AI-Finance, China}
  \country{}
}
\email{weirenjie@ai-finance.cn}

\author{Hongyan Pan}
\affiliation{%
  \institution{AI-Finance, China}
  \country{}
}
\email{panhongyan@ai-finance.cn}

\author{Cuiping Li}
\affiliation{%
  \institution{Renmin University of China, China}
  \country{}
}
\email{licuiping@ruc.edu.cn}

\author{Hong Chen}
\affiliation{%
  \institution{Renmin University of China, China}
  \country{}
}
\email{chong@ruc.edu.cn}

\renewcommand{\shortauthors}{Haoyang Li et al.}

\begin{abstract}
Language models have shown promising performance on the task of translating natural language questions into SQL queries (Text-to-SQL). However, most of the state-of-the-art (SOTA) approaches rely on powerful yet closed-source large language models (LLMs), such as ChatGPT and GPT-4, which may have the limitations of unclear model architectures, data privacy risks, and expensive inference overheads. To address the limitations, we introduce \model, a series of pre-trained language models with parameters ranging from 1B to 15B, specifically designed for the text-to-SQL task. \model is a fully open-source language model, which achieves superior accuracy with much smaller parameter sizes. This paper studies the research challenges in building \model. To enhance the SQL generation abilities of \model, we adopt an incremental pre-training approach using a specifically curated SQL-centric corpus. Based on this, we address the challenges of schema linking and rapid domain adaptation through strategic prompt construction and a bi-directional data augmentation technique. We conduct comprehensive evaluations on multiple datasets, including the widely used Spider benchmark, the newly released BIRD benchmark, robustness-diagnostic benchmarks such as Spider-DK, Spider-Syn, Spider-Realistic, and Dr.Spider, as well as two real-world datasets created for financial and academic applications. The experimental results show that our \model achieves new SOTA accuracy and robustness on nearly all challenging text-to-SQL benchmarks.


\end{abstract}

%
%


\received{20 February 2007}
\received[revised]{12 March 2009}
\received[accepted]{5 June 2009}

\maketitle

\section{Introduction}\label{sec:intro}
\begin{figure*}[t]
    \centering    \includegraphics[width=\textwidth]{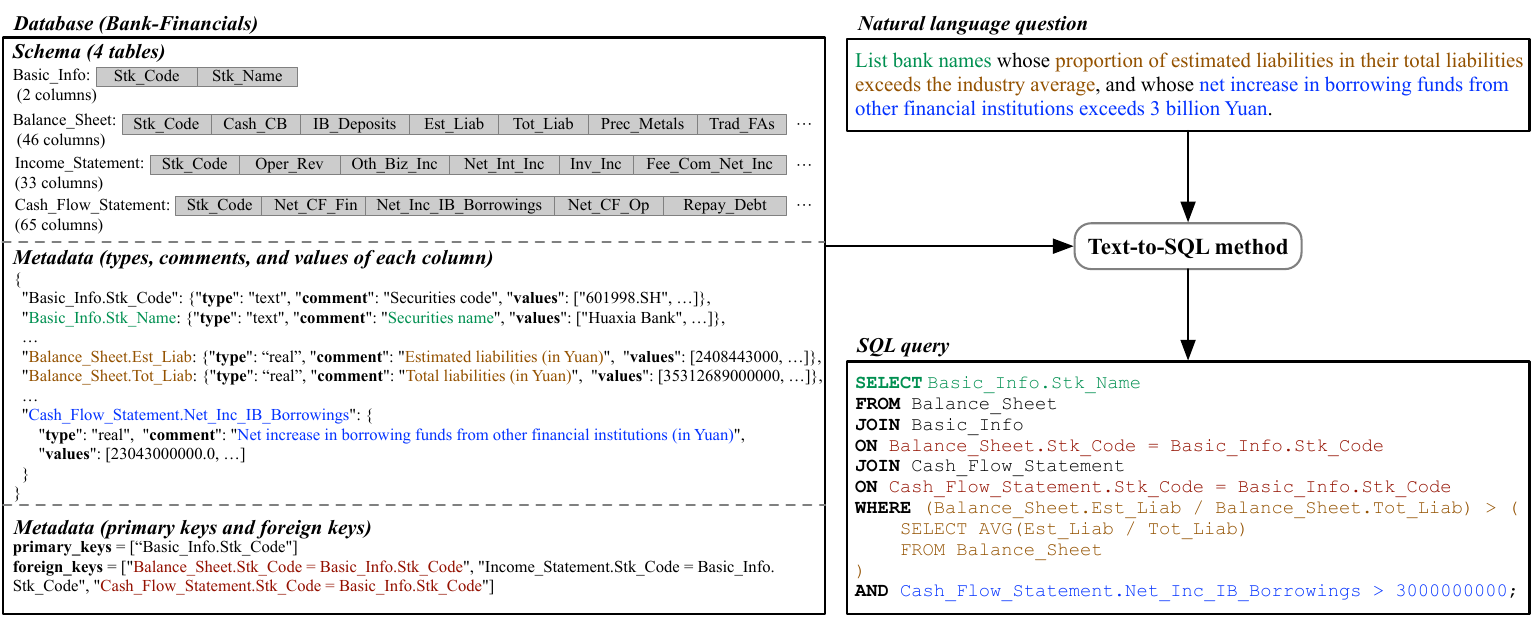}
	\caption{\label{fig:text2sql_example} An example of text-to-SQL in the finance domain.}
 \vspace{-1em}
\end{figure*}

The text-to-SQL task involves translating a user's natural language (NL) question into a corresponding and valid Structured Query Language (SQL) query that can be executed over a database.
Figure~\ref{fig:text2sql_example} illustrates how an NL question posed over a database (\emph{e.g.}, Bank Financial)
can be translated into an SQL query. 
Text-to-SQL enables users who may not be familiar with SQL or the structure of a database to interact with the database using natural language, and thus it has garnered increasing attention from both the database~\cite{fu2023@catsql, gu2023@fewshot, brunner2021@valuenet} and natural language processing communities~\cite{scholak2021@picard, li2023@resdsql}.

\begin{figure}[t]
    \centering
    \includegraphics[width=0.47\textwidth]{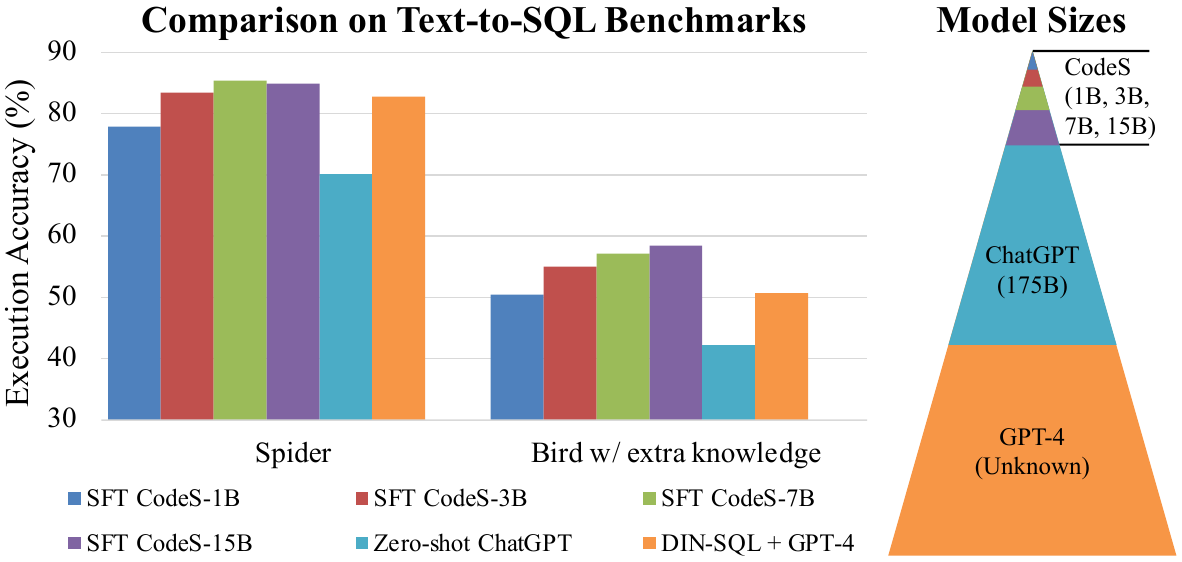}
	\caption{\label{fig:model_size_and_accuracy} Comparisons between \model and SOTA LLMs on two challenging text-to-SQL benchmarks, Spider~\cite{yu2018@spider} and BIRD~\cite{li2023@bird}. While 10x-100x smaller than the existing SOTA LLMs, \model achieves comparable or even superior accuracy.}
 \vspace{-2em}
\end{figure}

\stitle{State-of-the-Art: Strengths and Limitations.}
While traditional text-to-SQL utilizes the supervised fine-tuning approach (SFT)~\cite{wang2020@ratsql, scholak2021@picard, li2023@resdsql},
more recently, the paradigm has started to shift with the advent of large language models (LLMs) like GPT-4~\cite{openai2023@gpt4}, GPT-3.5~\cite{ouyang2022@instructgpt}, and PaLM-2~\cite{anil2023@palm2}. Instead of relying heavily on fine-tuning, LLMs have shown their capability in text-to-SQL using thoughtfully crafted prompts~\cite{chen2023@selfdebug, sun2023@sqlpalm, rajkumar2022@evaluating, liu2023@chatgptzeroshot, pourreza2023@dinsql, dong2023@c3}, which is known as ``prompt learning'' or ``in-context learning''~\cite{liu2023@pretrain}. 

However, most of the state-of-the-art (SOTA) approaches rely on \emph{closed-source} LLMs, such as DIN-SQL~\cite{pourreza2023@dinsql} based on GPT-4, SQL-PaLM~\cite{sun2023@sqlpalm} based on PaLM-2 and C3~\cite{dong2023@c3} built upon GPT-3.5. Although achieving promising text-to-SQL performance, these approaches may have the following limitations. (\textbf{L1}) Closed-source models hide their architectures and training/inference details, hindering purpose-specific continuous development tailored to specific applications. (\textbf{L2}) Invoking these models via APIs risks data privacy, as data must be sent to model providers. (\textbf{L3}) Most closed-source models have numerous parameters (e.g., 175B parameters for GPT-3.5), causing significant inference overheads, which are typically reflected by the monetary costs of calling APIs.

\stitle{Our proposals.}
This paper introduces \model, an \emph{open-source} language model that is designed for real-world text-to-SQL applications. 
\model is built upon StarCoder~\cite{li2023@starcoder}, a cutting-edge LLM designed specifically for code generation, with varying parameters ranging from 1B to 15B. Users can select an appropriately sized \model based on their computational resources to construct their text-to-SQL applications. As depicted in Figure~\ref{fig:model_size_and_accuracy}, compared with the SOTA text-to-SQL solutions, \model has the following advantages.
\begin{itemize}
    \item \textbf{Fully Open-source LLM.} \model, which is built upon StarCoder~\cite{li2023@starcoder}, is fully open-source and public to users.
    \item \textbf{New SOTA Results.} \model achieves the \textbf{new SOTA performance} on nearly all challenging text-to-SQL benchmarks, such as Spider~\cite{yu2018@spider} and BIRD~\cite{li2023@bird}.
    \item \textbf{Small Sizes.} \model is 10x-100x smaller than the existing SOTA LLMs, such as ChatGPT and GPT-4\footnote{It's worth noting that specific details regarding the scale of ChatGPT and GPT-4 have not been made public. Therefore, we follow the general assumption that ChatGPT is approximately equivalent in size to GPT-3-175B~\cite{kung2023@performance} and GPT-4 significantly exceeds the size of ChatGPT~\cite{nori2023@gpt4medical, openai2023@gpt4}.}.
\end{itemize}

\stitle{Challenges and Solutions.}
We outline the main technical challenges in developing \model and explain our solutions as follows.

\stitle{C1: How to enable small-sized language models with complex text-to-SQL reasoning capacity?} 
Directly using pre-trained language models, such as LLaMA-2~\cite{touvron2023@llama2} and StarCoder~\cite{li2023@starcoder}, faces challenges in text-to-SQL, mainly because SQL-related content typically constitutes only a tiny fraction of their entire pre-training corpus. Subsequently, the data bias could potentially hinder the text-to-SQL capability, as language models during the pre-training phase aim to fit the distribution of the entire corpus, rather than just the distribution of SQL-related data.
Moreover, compared with ChatGPT or GPT-4, the small-sized language models possess limited reasoning capacity, resulting in insufficient capabilities on text-to-SQL. To address the challenge, we propose an incremental pre-training approach utilizing a vast curated dataset relevant to the text-to-SQL task. Specifically, we collected 21.5GB of data, consisting of SQL-related data (11GB), NL-to-code data (6GB), and NL-related data (4.5GB), with the aim of enhancing both SQL generation capabilities and natural language comprehension. By incrementally pre-training StarCoder~\cite{li2023@starcoder} on this corpus, we created a series of {\smodel} models, with varying parameters ranging from 1B to 15B.

\stitle{C2: How to generate high-quality prompts for text-to-SQL to alleviate the difficulty of schema linking?}
Schema linking is crucial for text-to-SQL translation, ensuring models map input questions to specific database elements~\cite{lei2020@schemalinking}. Yet, issues emerge with numerous tables, wide tables with numerous columns, ambiguous table/column names, and large tables with vast values. 
%
To address the challenge, we use a schema filtering strategy for numerous tables and wide tables, retaining only relevant tables and columns based on the user's query from the database, ensuring the schema doesn't exceed the model's context length.
For ambiguous names, like abbreviations, we tie comments to these names, aiding models in understanding context.
For large tables, a ``coarse-to-fine'' approach is adopted: we initially filter values using the BM25 index based on the question and further refine them with the longest common substring algorithm.
Using these techniques, we frame prompts for the \smodel model, enhancing schema linking and boosting text-to-SQL performance for complex databases.

\stitle{C3: How to adaptively transfer to databases within a new domain?} 
In real-world applications, we aim for the \smodel model to adapt across various domains. A significant obstacle, however, is the lack of specific (question, SQL) pairs for fine-tuning.
%
To address this challenge, we employ a bi-directional data augmentation technique. Firstly, we collect a few genuine user queries, manually annotate corresponding SQL queries, and leverage GPT-3.5 to produce a wider set of (question, SQL) pairs, ensuring user-oriented authenticity.
On the other hand, we utilize SQL templates and their question templates from text-to-SQL benchmarks. By plugging in tables, columns, and values from the databases
of new domains, we generate a diverse set of (question, SQL) pairs. This templating approach aids \smodel's adaptability to new distributions.
In essence, our crafted training dataset combines real examples with structured templates, guaranteeing both authenticity and broad applicability.

\stitle{Evaluation.}
We evaluate the created \smodel on two challenging text-to-SQL benchmarks: Spider~\cite{yu2018@spider} and BIRD~\cite{li2023@bird}. While Spider has long been a standard for text-to-SQL, BIRD offers more complex questions, ambiguous schema, and large and dirty database values. The leading text-to-SQL method, DIN-SQL+GPT-4, manages only around 56\% on its test set. Beyond the conventional Spider benchmark, we also assess \smodel against Spider's four distinct variants: Spider-DK~\cite{gan2021@spider-dk}, Spider-Syn~\cite{gan2021@spider-syn}, Spider-Realistic~\cite{deng2021@spider-realistic}, and Dr.Spider~\cite{chang2023@drspider}. These span a total of 20 test sets and are tailored to probe model resilience, especially in scenarios where test data distributions differ from training data distributions. To investigate the effect of our bi-directional data augmentation approach in rapidly adapting to new domains, we sourced databases from both the academic and finance domains. Given that both databases had insufficient training data for effective fine-tuning, we augmented the training data, fine-tuned our model, and then evaluated its performance on the respective test sets.

Our contributions are summarized as follows:
\begin{itemize}
    \item We introduce \model, a series of language models ranging from 1B to 15B parameters, designed specifically for SQL generation. This innovation is underpinned by an incremental pre-training technique and a meticulously curated pre-training corpus, comprising SQL-related, NL-related, and NL-to-code data. This approach marks a significant advancement in language models tailored for text-to-SQL applications.
    \item We enhance \model's performance using a comprehensive database prompt. Additionally, to facilitate its adaptation to new domains, we introduce a bi-directional data augmentation approach with limited annotation overhead.
    \item Through extensive evaluations on multiple text-to-SQL benchmarks, we demonstrate that (1) \smodel surpasses other notable open-source pre-trained language models in SQL generation capability; (2) When fine-tuned, \smodel achieves new SOTA accuracy and robustness on almost all challenging text-to-SQL benchmarks.
    \item We have open-sourced our code, models, and data on GitHub\footnote{\url{https://github.com/RUCKBReasoning/codes}} to foster further research, adoption, and innovation in the text-to-SQL domain within the community.
\end{itemize}
\section{Related Work}\label{sec:related_work}
In our survey, we cover supervised fine-tuning and prompting-based methods for text-to-SQL. Furthermore, we explore existing code language models because text-to-SQL can be viewed as a sub-task of code generation. Additionally, we examine various schema linking and data augmentation techniques that have been proposed to enhance text-to-SQL methodologies.

\stitle{Supervised Fine-Tuning-Based Text-to-SQL.}
Before the era of LLMs, the mainstream approach in text-to-SQL is fine-tuning an ``encoder-decoder'' neural network model. Significant efforts have been made to enhance the representation capability of the encoder that encodes both the question and the database by incorporating graph structural information that exists between query tokens, tables, and columns using graph-relational neural networks~\cite{wang2020@ratsql, cao2021@lgesql}. Some other efforts focus on injecting SQL grammar into the decoder, which constrains the output space of the decoder, ensuring the generation of syntactically correct SQL queries~\cite{yin2017@a, scholak2021@picard, fu2023@catsql}. With the advancement of language models, there has been an increasing trend in formatting text-to-SQL as a sequence-to-sequence task~\cite{scholak2021@picard, shaw2021@compositional, li2023@graphix, li2023@resdsql, dou2022@unisar, gu2023@fewshot}, where the input sequence consists of a natural language question and flattened database information including tables, columns, foreign keys, etc., and the output sequence is the target SQL query. Then, sequence-to-sequence language models such as T5~\cite{raffel2020@t5}, and BART~\cite{lewis2020@bart} are fine-tuned on these input-output sentence pairs, enabling them to generate SQL queries from the provided input. Inspired by the remarkable achievements of pre-training techniques~\cite{devlin2019@bert, radford2018@gpt, raffel2020@t5}, a series of studies~\cite{yu2021@grappa, shi2021@gap, herzig2020@tapas, yin2020tabert, deng2021@strug} have explored pre-training language models using extensive database-related data and various pre-training objectives. However, in contrast to our \model, their primary goal isn't directly enhancing the SQL generation capability of language models. Instead, they focus on enhancing the encoder's ability to better represent the question and the database schema. Then, these pre-trained encoders are integrated into the ``encoder-decoder'' models.

\stitle{Prompting-Based Text-to-SQL.}
The advent of LLMs, such as GPT-3~\cite{brown2020@gpt3}, Codex~\cite{chen2021@codex}, PaLM~\cite{chowdhery2022@palm}, LLaMA~\cite{touvron2023@llama}, StarCoder~\cite{li2023@starcoder}, has brought about a revolutionary transformation in the field of NLP, achieving remarkable progress on various complex tasks that require reasoning abilities without fine-tuning any parameters~\cite{chen2023@selfdebug,liu2023@chatgptzeroshot, pourreza2023@dinsql}. For text-to-SQL, by leveraging a few text-to-SQL demonstrations as the few-shot prompt, SQL-PaLM~\cite{sun2023@sqlpalm} (based on PaLM-2) and Self-Debugging~\cite{chen2023@selfdebug} (based on Codex) successfully achieve the state-of-the-art (SOTA) performance on Spider. In addition, inspired by chain-of-thought reasoning~\cite{wei2022@cot}, designing effective prompts to stimulate the text-to-SQL capability of LLMs has become a hot research topic. DIN-SQL~\cite{pourreza2023@dinsql} (based on GPT-4) breaks down the text-to-SQL task into several simpler sub-tasks, including schema linking, query classification \& decomposition, and SQL generation. C3~\cite{dong2023@c3} (based on ChatGPT), by providing appropriate instructions to ChatGPT, makes it become an experienced zero-shot text-to-SQL parser. Although these prompting-based methods have achieved SOTA performance on text-to-SQL benchmarks, as we analyzed in Section~\ref{sec:intro}, it is challenging to implement them in real-world applications due to the significant costs and potential data privacy concerns associated with using these models' APIs.

\stitle{Code Language Models.}
Over the past few years, there has been a growing interest in leveraging language models for coding-related tasks such as code understanding and generation~\cite{chen2021@codex, nijkamp2023@codegen, nijkamp2023@codegen2, li2023@starcoder}. Existing code language models are often pre-trained on a diverse mix of programming languages, such as C, C++, Python, Java, C\#, and SQL. This broad-spectrum training data can result in suboptimal performance for small-scale models on a specific programming language (\emph{e.g.}, SQL) due to their constrained representation ability. To address this issue, we develop \model – a collection of open-source generative language models, uniquely optimized with an emphasis on a mixture of SQL-centric data.

\stitle{Schema Linking.} Schema linking plays a crucial role in text-to-SQL processes, aiming to identify referenced database schemas (such as tables and columns) and database values within natural language questions~\cite{lei2020@schemalinking}. There are primarily two strategies for schema linking: string matching-based~\cite{wang2020@ratsql, guo2019@towards, dou2022@unisar, brunner2021@valuenet} and neural network-based~\cite{bogin2019@global, li2023@resdsql, gu2023@zeronl2sql}. The string matching-based approach, simple yet effective, identifies schemas and values related to a question through direct string matching. However, this approach has limitations in certain scenarios, such as dealing with synonyms. To address these challenges, neural network-based methods have been developed. These methods are designed to assess the relevance of schemas and values at a semantic level. Once the schema linking results are obtained, for example, the matching degrees for all tables and columns, many techniques~\cite{brunner2021@valuenet, dou2022@unisar, wang2020@ratsql} incorporate these results as additional input for the text-to-SQL model. Different from them, RESDSQL~\cite{li2023@resdsql} utilizes these results for schema pruning, retaining only the schemas most relevant to the input of the subsequent neural network, thus reducing the length of the input to the LLMs.

\stitle{Data Augmentation for Text-to-SQL.} Recent times have seen a growing interest in synthesizing data for text-to-SQL. The aim is to automatically generate (question, SQL) pairs that are relevant to a given database. Many current methods~\cite{hu2023@importance, wang2021@learning, wu2021@data, zhang2023@sciencebenchmark} employ a SQL-to-question synthesis pipeline. This process typically involves first auto-generating SQL queries according to a database, and then translating these queries into natural language questions using a sequence-to-sequence model. However, a notable drawback of these methods is that the synthesized natural language questions often starkly differ from actual users. To bridge this gap, we propose a novel bi-directional data augmentation strategy. This approach not only leverages SQL-to-question synthesis but also incorporates question-to-SQL synthesis, more accurately generating the variety of questions real-world users might ask.
\begin{figure*}[t]
    \centering
    \includegraphics[width=0.83\textwidth]{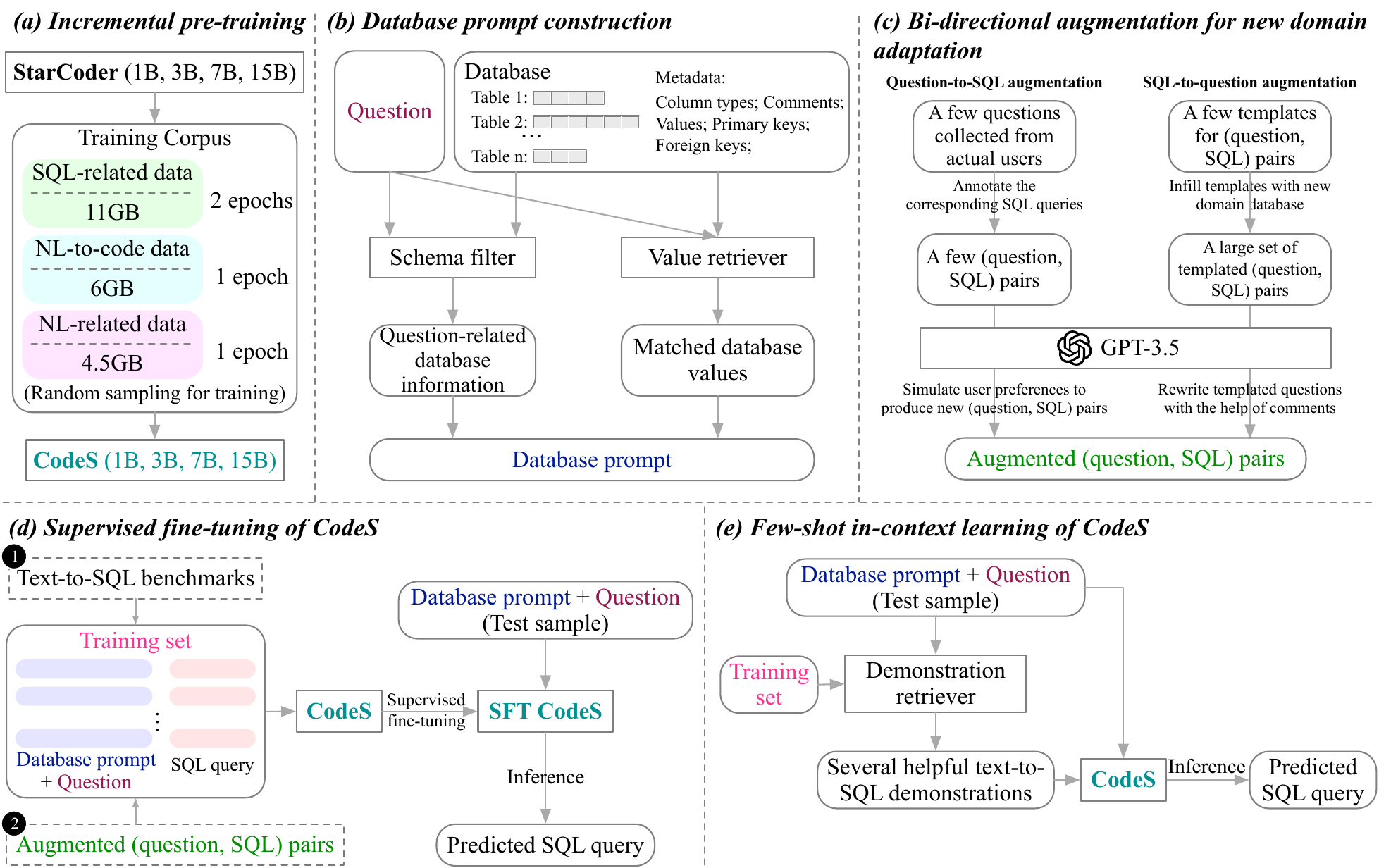}
	\caption{\label{fig:overview} Illustration of the comprehensive framework which encompasses:
(a) \smodel that is incrementally pre-trained on top of StarCoder using our specially curated SQL-focused dataset.
(b) Our unique method for database prompt construction.
(c) The proposed bi-directional data augmentation technique for adapting to new domains.
\smodel can be employed in two distinct manners:
(d) Inferring after a supervised fine-tuning of \smodel on a training dataset, sourced from text-to-SQL benchmarks along with our enriched (question, SQL) pairs.
(e) Direct inference through few-shot in-context learning on \model.}
\end{figure*}

\section{Preliminaries}\label{sec:preliminaries}

\stitle{Text-to-SQL Task.}
The objective of text-to-SQL is to generate a SQL query $S$ based on a natural language question $Q$ and a database $D$, such that the SQL query can be executed to answer the question:
\begin{equation}
\begin{aligned}
    S = Parser(Q, D), \\
\end{aligned}
\end{equation}

\noindent where the $Parser()$ is designed to interpret the provided $Q$ using $D$ and produce $S$. $D$ contains database schema (\emph{i.e.}, tables and columns) and database metadata, which contains column types, comments, column values, primary keys, and foreign key relations. An illustrative text-to-SQL example is presented in Figure~\ref{fig:text2sql_example}.

\stitle{Pre-trained Language Models.}
Language models, primarily based on the Transformer architecture~\cite{vaswani2017@transformer}, excel at text understanding and generation tasks. They typically undergo an initial pre-training phase on extensive text data using unsupervised learning objectives. Two prominent unsupervised learning paradigms are ``language modeling'' and  ``masked language modeling''. In the former, models like GPT~\cite{radford2018@gpt}, PaLM~\cite{chowdhery2022@palm}, and LLaMa~\cite{touvron2023@llama} predict the next word from a given context. In masked language modeling, specific words or spans within the input are randomly masked, and the task is to reconstruct the masked segments based on the unmasked context. Representative pre-trained language models using this approach include BERT~\cite{yin2020tabert}, RoBERTa~\cite{liu2019@roberta}, and T5~\cite{raffel2020@t5}. 

\stitle{Supervised Fine-Tuning and In-context Learning.}
Pre-trained language models possess extensive linguistic knowledge, yet specific tasks often demand unique language patterns or domain expertise. To address this,
supervised fine-tuning (SFT) involves further training the model on task-specific labeled data, leveraging its initial pre-training and insights gained from the new dataset.
In contrast to SFT, the concept of ``in-context learning''~\cite{liu2023@pretrain, wei2022@cot} enables language models to perform new tasks by simply providing appropriate prompts in the input, without the need for SFT. However, the effectiveness of in-context learning depends heavily on the quality of the prompts and the language model itself.

\section{Overview}\label{sec:overview}
As illustrated in Figure~\ref{fig:overview}, we introduce three components to solve the challenges in developing a compact but powerful text-to-SQL model and show the flexible usage of \model.

\stitle{Incremental Pre-training (Figure~\ref{fig:overview}(a) and Section~\ref{sec:pretraining}).} To improve existing language models' SQL generation and natural language understanding capabilities, we first collect a new corpus consisting of 11GB of SQL-related data, 6GB of NL-to-code data, and 4.5 GB of NL-related data from diverse sources. Then, based on StarCoder, we perform incremental pre-training using the SQL-centric corpus and obtain our pre-trained language model \model, which is available in 4 scales: 1B, 3B, 7B, and 15B.

\stitle{Database Prompt Construction (Figure~\ref{fig:overview}(b) and Section~\ref{sec:db_prompt}).} We present a comprehensive database prompt construction approach to generate high-quality database prompts. The strategy mainly contains a schema filter and a value retriever. The schema filter is to eliminate irrelevant tables and columns based on the given question. The value retriever is tailored to extract potentially useful database values that align with the question. In addition to table and column names, we also incorporate various metadata, including data types, comments, representative column values, and information on primary and foreign keys. This comprehensive inclusion serves to provide a richer context for text-to-SQL models.

\stitle{New Domain Adaptation (Figure~\ref{fig:overview}(c) and Section~\ref{sec:adaption}).} We present a bi-directional data augmentation method to produce a vast set of (question, SQL) pairs for new domain databases. In the question-to-SQL direction, we start with a few real-world questions, label their SQL counterparts, and use GPT-3.5 to expand the dataset.  
For the SQL-to-question approach, we extract templates from existing text-to-SQL benchmarks, infill the templates with the schema of the new database, and refine the questions with GPT-3.5. 
This bidirectional strategy creates a robust training set with minimal annotation, easing model fine-tuning for the new domain.

\stitle{The Usage of \smodel (Figure~\ref{fig:overview}(d), (e) and Section~\ref{sec:usage}).} In cases where abundant training data is accessible, we can directly fine-tune the parameters of the \model, tailoring it to closely match the distribution of labeled data (Figure~\ref{fig:overview}(d)). Conversely, in scenarios with limited training data, we can utilize the \model's in-context learning ability without any fine-tuning by only offering a few text-to-SQL demonstrations (Figure~\ref{fig:overview}(e)). A demonstration retriever is introduced to extract valuable demonstrations by concurrently considering both question similarity and question-pattern similarity. 

\stitle{Complexity Discussion.}
In discussing the complexities of our proposed solution, which is composed of various components, it's essential to distinguish between offline and online processes. Specifically, the incremental pre-training (Section~\ref{sec:pretraining}) and new domain adaptation (Section~\ref{sec:adaption}) are conducted offline, meaning they are executed only once. Conversely, the database prompt construction strategy (Section~\ref{sec:db_prompt}) is an online process, as it must respond to each user's query during inference. The complexity of prompt construction arises from two main components: the schema filter and the value retriever. The schema filter employs a compact neural network for schema classification, which achieves fast inference speeds. On the other hand, the value retriever's efficiency is significantly enhanced by integrating BM25 indexing, leading to a noticeable acceleration in its online processing speed.

\section{Incremental Pre-Training}\label{sec:pretraining}

\subsection{Pre-training Corpus}\label{sec:corpus}
We enhance the text-to-SQL model's capabilities in SQL generation and natural language understanding through the collection of datasets from three key dimensions: SQL-related data, natural language-related data, and natural language-to-code data.

\stitle{SQL-related data [11GB].}
To further enhance the SQL generation capability of language models, we employ the SQL segment from StarCoder's pre-training corpus~\cite{li2023@starcoder}.

\stitle{NL-related data [4.5GB].}
To bolster the capability in natural language comprehension, we sample 4.5GB of high-quality dialog data from three sources: (1) Alpaca-cleaned\footnote{\url{https://huggingface.co/datasets/yahma/alpaca-cleaned}} is designed for developing an instruction-following language model. This dataset is constructed using the self-instruct technique~\cite{wang2023@selfinstruct}, aided by OpenAI's text-davinci-003 model. (2) Unnatural-instructions~\cite{honovich2023@unnatural} is also a large instruction-following dataset collected with almost no human labor. Both alpaca-cleaned and unnatural-instructions datasets can be characterized as single-turn dialogues. (3) UltraChat~\cite{ding2023@ultrachat} is a multi-turn dialogue dataset, produced by iteratively invoking two distinct GPT-3.5 APIs.

\stitle{NL-to-code data [6GB].}
To bridge the gap between natural language questions and SQL queries, we incorporate four types of NL-to-code datasets into our pre-training corpus: (1) CoNaLa~\cite{yin2018@conala} and StaQC~\cite{yao2018@staqc} are derived automatically from Stack Overflow, encompasses many NL-to-Python and NL-to-SQL pairs. (2) CodeAlpaca-20k\footnote{\url{https://huggingface.co/datasets/sahil2801/CodeAlpaca-20k}} encompasses a wealth of instruction-following data related to code, being created using the self-instruct methodology~\cite{wang2023@selfinstruct}. (3) Jupyter-structured-clean-dedup, a subset of the StarCoder's pre-training corpus, comprises a vast collection of structured Jupyter notebooks containing both code and accompanying natural language explanations. (4) Unlike the previously mentioned datasets, NL-SQL-458K is a brand-new dataset crafted by us, containing a vast number of NL-SQL pairs. Specifically, we start by using regular expressions to extract all ``\texttt{SELECT}'' queries from three extensive open-source corpora: The Pile~\cite{gao2021@thepile}, The Stack~\cite{kocetkov2022@thestack}, and GitHub Code\footnote{\url{https://huggingface.co/datasets/codeparrot/github-code}}. We then filter out queries with syntax errors, resulting in 458K SQL queries. To generate corresponding natural language questions for each SQL query, we employ GPT-3.5, using the prompts of eight paired (SQL, question) demonstrations.

\begin{table}[!t]
    \centering
    \caption{Details about architectures of \model. ``Shared'': remains consistent across all model versions; ``Distinct'': varies among different model versions.}
    \vspace{-1em}
    \scalebox{0.95}{
    \begin{tabular}{@{}l|cc@{}}
        \toprule
        \textbf{Type} & \textbf{Hyper-parameter} & \textbf{Value} \\
        \midrule
        \multirow{5}{*}{Shared}    & Transformer architecture & Decoder-only \\
                                            & Position embedding & Learned absolute embeddings \\
                                            & Attention type & Multi-query \\
                                            & FlashAttention-2 & Enable \\
                                            & Vocabulary size & 49,152 \\
        \midrule
        \multirow{5}{*}{Distinct}    & \#Parameters & 1B/3B/7B/15B \\
                                            & Maximum context length & 8,192/8,192/8,192/6,144 \\
                                            & Transformer's hidden size & 2,048/2,816/4,096/6,144 \\
                                            & Feed-forward hidden size & 8,192/11,264/16,384/24,576 \\
                                            & \#Attention heads & 16/22/32/48 \\
                                            & \#Transformer blocks & 24/36/42/40 \\
        \bottomrule
    \end{tabular}
    }
    \label{tab:hyperparams}
    \vspace{-1em}
\end{table}

\subsection{Pre-Training Details}\label{sec:pt_details}
\smodel is built upon StarCoder, a series of open-source code language models pre-trained on a mixture of over 80 programming languages (such as C, Python, Java, PHP, SQL, and others), Jupyter notebooks, GitHub issues, and Git commits. To develop \model, we incrementally pre-train StarCoder for two epochs on SQL-related data and one epoch each on NL-related and NL-to-code data. This mixed training in natural language and code offers benefits for a wide range of tasks in both domains~\cite{nijkamp2023@codegen2}. Specifically, \model-15B is based on StarCoder-15B, while \model-(1B, 3B, 7B) are derived from the respective StarCoderBase-(1B, 3B, 7B). Then we optimize the language modeling objective that is widely used in prior pre-trained language models like GPT~\cite{radford2018@gpt} and LLaMA~\cite{touvron2023@llama}. Specifically, given a sequence $x$ consisting of $n$ tokens, denoted as ${t_0, t_1, t_2, ..., t_{n-1}}$, our objective is to maximize the likelihood of the entire sequence. This is achieved by calculating the product of the conditional probabilities for each token:

\begin{equation}
    p(x) = \prod_{i=1}^{n-1} p(t_i|t_1, t_2, ... t_{i-1}).
\end{equation}

To optimize the objective, we employ the AdamW optimizer~\cite{loshchilov2019@adamw} with parameters set to $\beta_{1}=0.9$, $\beta_{2}=0.95$, and $\epsilon=10^{-8}$. The learning rate is configured at $5e^{-5}$, accompanied by a weight decay of 0.1. Our learning rate scheduler follows a cosine decay without any warm-up steps, and we set the final learning rate at a tenth of its initial value. The training process uses a large batch size comprising 4M tokens, and to ensure stability, we apply gradient clipping with a clipping value of 1.0. We leverage the DeepSpeed Zero-3 framework~\cite{rajbhandari2020@deepspeedzero}, employing BF16 mixed precision, to optimize GPU memory consumption during pre-training. More details about model architectures can be found in Table~\ref{tab:hyperparams}. We integrate the FlashAttention-2~\cite{dao2023@flashattention2} into \smodel, enhancing its capability to handle extended context lengths. However, due to GPU memory constraints, \model-(1B, 3B, 7B) can handle a maximum context length of 8,192, whereas \model-15B is limited to 6,144. In practice, incremental pre-training over our collected corpus takes approximately 1.5, 3, 8, 16 days for \model-(1B, 3B,7B, 15B).

\section{Database Prompt Construction}\label{sec:db_prompt}
Beyond model advancements, building effective database prompts is crucial for the text-to-SQL task. High-quality prompts furnish the language model with valuable insights, enabling it to generate precise SQL queries more efficiently. To craft these superior database prompts, we employ two key strategies: a schema filter and a value retriever, while also incorporating crucial database metadata. The pseudo-code detailing our prompt construction process is presented in Algorithm~\ref{alg:db_prompt}.

\subsection{Schema Filter}\label{sec:schema_filter}
In real-world scenarios, databases often encompass a vast array of tables and columns, resulting in extremely long database prompts. When these prompts surpass the language model's maximum context length, truncation becomes necessary. However, this process may omit vital tables or columns that are necessary for generating the target SQL query. Hence, it's imperative to adopt a method that minimizes the database prompt length without sacrificing critical schema information. Following~\cite{li2023@resdsql}, we employ a schema filter to retain the most relevant tables and columns within the database for each text-to-SQL sample. Concretely, given a database and a question, we develop a schema item classifier, which is trained to predict relevance scores for the tables and columns according to the user's question. Utilizing these scores, we retain the top $top_{k1}$ tables and, for each of these tables, the top $top_{k2}$ columns. Then, for text-to-SQL samples in the training set, since the ground truth SQL query is available, we could directly identify the used tables and columns. Aiming for consistency in distribution between test and training data, if the number of the used tables falls below $top_{k1}$, we incorporate randomly selected, unused tables from the database as padding. A similar procedure is adopted for columns, ensuring each retained table contains $top_{k2}$ columns. The integration of the schema filter not only reduces the length of the database prompts but also alleviates the schema-linking pressure for the model.

\subsection{Value Retriever}\label{sec:value_retriever}
Retrieving values from the database that align with the question can help the language model perform better schema linking when generating predicates. For a natural language question in the BIRD benchmark ``\texttt{How many clients opened their accounts in Jesenik branch were women?}'', a comparison against the database's values reveals that the ``\texttt{a2}'' column in the ``\texttt{district}'' table holds the value ``\texttt{Jesenik}''. Subsequently, integrating the information ``\texttt{district.a2 = 'Jesenik'}'' into the database prompt can guide the model in producing accurate predicates for the SQL query. Prior work~\cite{scholak2021@picard, li2023@resdsql, li2023@graphix} consistently uses the longest common substring (LCS) algorithm to calculate the matching degree of a database value to a question. However, the time complexity of this algorithm is $O(f*u)$, where $f$ and $u$ denote the lengths of the two input strings. In many scenarios where the database contains a vast amount of values (for instance, the Donor database in the BIRD benchmark, which encompasses approximately 116.5 million valid values), using LCS to calculate the matching degree for every database value becomes excessively time-consuming. To tackle this, we propose a ``coarse-to-fine'' matching approach. The essence of this method lies in utilizing indexes for a fast yet coarse-grained initial search, followed by a fine-grained matching process using the LCS algorithm. Specifically, we use Lucene\footnote{\url{https://github.com/castorini/pyserini}} to build the BM25 index for all values stored in each database. When a user's question is received, the indexes first extract hundreds of potentially relevant values from the whole database based on the question. Then, the LCS method is employed to calculate the degree of match between the question and these potentially relevant values to find the most relevant values. The integration of the BM25 index significantly enhances the value retrieval speed for the extensive database by drastically reducing the number of LCS algorithm invocations from potentially millions to just hundreds.

\subsection{Database Metadata}\label{sec:metadata}
In our database prompt, we additionally incorporate certain metadata that is valuable for text-to-SQL:

\stitle{(1) Column Data Types:} The data type of a column dictates its validation rules and permissible operations. For instance, numeric types like \texttt{INTEGER} and \texttt{REAL} support arithmetic operations, whereas string types don't. If certain data is stored as a string type, the \texttt{CAST} function must be used in the SQL query before performing arithmetic operations on it.

\stitle{(2) Comments:} The ambiguities in database schemas are prevalent in real-world databases. Table~\ref{tab:comments} shows examples of ambiguous columns from the BIRD benchmark. Ambiguous schemas can lead models to select the wrong tables or columns in their generated SQL queries, as language models typically use semantic similarity for schema linking. Fortunately, databases usually provide informative comments for ambiguous schema. We incorporate these comments into both the input of the schema item classifier and the database prompts to facilitate the LLM to perform accurate schema linking.

\begin{table}[!t]
    \centering
    \caption{Examples of ambiguous columns in BIRD dataset.}
    \vspace{-1em}
    \begin{tabular}{lll}
        \toprule
        Database & Column name & Comment\\
        \midrule
        language\_corpus & w1st & word id of the first word \\
        hockey & rotl & road overtime loses \\
        ice\_hockey\_draft & pim & penalty minutes \\
        \bottomrule
    \end{tabular}
    \label{tab:comments}
    \vspace{-1em}
\end{table}

\stitle{(3) Representative Database Values:} In addition to column names, representative column values are beneficial. For instance, to generate predicates such as ``\texttt{client.gender = 'F'}'', it's imperative to inform the language model that the ``\texttt{gender}'' column offers two values: 'M' and 'F'. Similarly, for predicates like ``\texttt{SUBSTR (hiredate, 1, 4) = '2009'}'', the model should be aware of the ``hiredate'' column's specific format, ``YYYY-MM-DD''. Apparently, question-matched values don't always address these nuances. To address this, we extract representative values for each column. Drawing inspiration from~\cite{chang2023@how}, we employ the SQL query ``\texttt{SELECT DISTINCT \{COLUMN\} FROM \{TABLE\} WHERE \{COLUMN\} IS NOT NULL LIMIT 2}'' to capture two distinct representative non-null values from each column. By offering these insightful values, the language model is better positioned to produce precise and context-relevant SQL queries.

\stitle{(4) Primary and Foreign Keys:} The primary key uniquely identifies each row in a table, while the foreign key creates associations between tables. Incorporating primary and foreign key information can guide the model to deduce the appropriate join path, ensuring accurate \texttt{JOIN ON} clause generation. In practice, we use a unique identifier for every primary key column and represent each foreign key as ``\texttt{\{TABLE1\}.\{COLUMN1\} = \{TABLE2\}.\{COLUMN2\}}'' within the database prompt.

\begin{algorithm}[]
    \caption{Database Prompt Construction}
    \small
    \label{alg:db_prompt}
    \LinesNumbered
    \KwIn{user question $Q$, schema item classifier model $M_{cls}$, database schema $D_{schema}$, database metadata $D_{meta}$, pre-built BM25 index $I$, maximum table and column number $top_{k1}, top_{k2}$}
    \KwOut{Database prompt $Prompt_{db}$}
        \tcp{Apply schema filter}
        $computeRelevantScores(Q, D_{schema}, M_{cls}) \rightarrow Scores$\;
        $filterDBInfo(Scores, D_{schema}, D_{meta}, top_{k1}, top_{k2}) \rightarrow D_{schema}^{r}, D_{meta}^{r}$\;
        \tcp{Apply value retriever} 
        $BM25Matching(Q, I) \rightarrow V_{coarse-grained}$\;
        $LCS(Q, V_{coarse-grained}) \rightarrow V_{fine-grained}$\;
        \tcp{Serialization and concatenation}
        $Serialize(D_{schema}^{r}, D_{meta}^{r}) \rightarrow S_{db}$\;
        $Serialize(V_{fine-grained}) \rightarrow S_{value}$\;
        $ConcatSequence(S_{db}, S_{value}) \rightarrow Prompt_{db}$\;
        \textbf{return} $Prompt_{db}$
\end{algorithm}

\begin{figure}[t]
    \centering
    \includegraphics[width=0.46\textwidth]{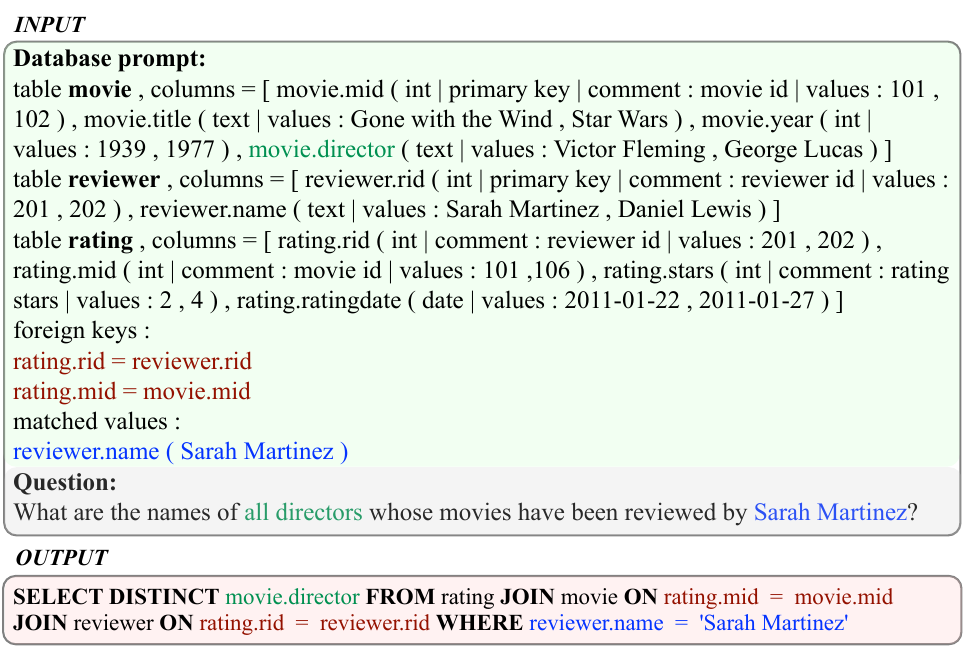}
	\caption{\label{fig:text2sql_prompt} A text-to-SQL sample in Spider's training set, consisting of a triplet of <database prompt, question, SQL query>. The database prompt is crafted by our proposed method.}
    \vspace{-1em}
\end{figure}

In Figure~\ref{fig:text2sql_prompt}, we show a training sample from the Spider benchmark, consisting of paired input and output sequences. This database prompt is crafted using our proposed strategy. As seen, based on the user's question, a relevant database value, \texttt{Sarah Martinez}, is extracted from the \texttt{reviewer.name} column. Then, the displayed primary and foreign keys further guide the language model in generating the \texttt{JOIN ON} clauses.

\section{New Domain Adaption}\label{sec:adaption}
In real-world scenarios, people usually have their own databases from various new domains such as finance, genes, biology, academia, and more. However, developing a text-to-SQL model on these databases is challenging because of the lack of labeled training data. In this section, we propose a bi-directional data augmentation technique to automatically generate a large set of authentic and general (question, SQL) pairs with minimal annotation costs.

\begin{figure}[t]
    \centering
    \includegraphics[width=0.45\textwidth]{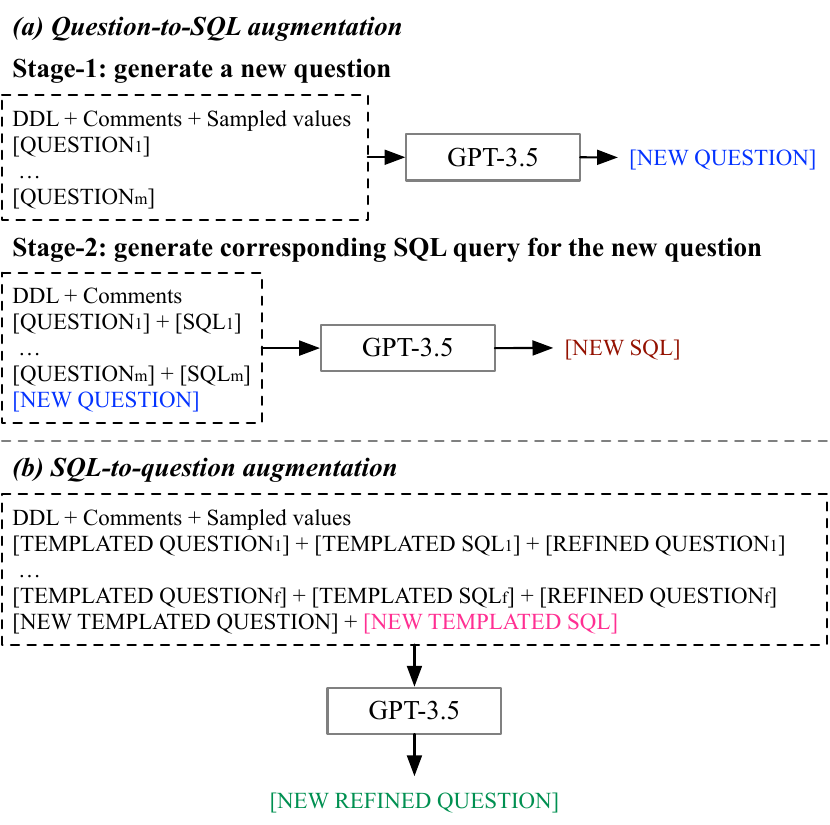}
    \vspace{-1em}
	\caption{\label{fig:augmentation} Prompt formats used in the bi-directional data augmentation. DDL stands for data definition language.
 }
 \vspace{-1em}
\end{figure}

\stitle{Question-to-SQL Augmentation}. This augmentation direction seeks to produce genuine (question, SQL) pairs aligned with user preferences. Specifically, we first gather a few authentic and representative natural language questions from real-world text-to-SQL users. These questions embody the most common inquiries users have for specific databases. Then, we manually annotate their corresponding SQL queries to obtain a few high-quality (question, SQL) pairs. Given that frequently asked questions are typically few in number, the annotation effort is relatively minimal. Furthermore, given the constrained volume of annotated data, it's insufficient for directly fine-tuning language models. To address this, we introduce a two-stage prompting approach: Initially, we prompt GPT-3.5 to generate potential questions, drawing inspiration from the real questions we've gathered, effectively capturing user intents. Then, we employ GPT-3.5 to provide corresponding SQL queries for these synthesized questions. Figure~\ref{fig:augmentation}(a) shows the prompts used during the two-stage process. Here, [QUESTION${}{i}$] and [SQL${}{i}$] denote a pair of manually labeled data, with $m$ indicating the total number of such pairs. To guarantee the diversity of the newly generated questions, we shuffle the sequence of user-providing questions and employ a high-temperature hyperparameter for each generation. Lastly, [NEW QUESTION] and [NEW SQL] represent a new data pair that mirrors user preferences, ensuring its authenticity.

\stitle{SQL-to-Question Augmentation.} Inspired by SyntaxSQLNet~\cite{yu2018@syntaxsqlnet}, we explore another augmentation method that generates generic (question, SQL) pairs using a set of universal templates. Specifically, this paper uses (question, SQL) templates derived from Spider, a widely-recognized text-to-SQL benchmark, encompassing 75 common SQL templates. Examples include a templated question: ``\texttt{Return the lowest \{COLUMN\} of \{TABLE\}}'', and its corresponding templated SQL: ``\texttt{SELECT \{COLUMN\} FROM \{TABLE\} GROUP BY \{COLUMN\} ORDER BY COUNT (*) ASC LIMIT 1}''. Given the versatility of natural language, a single SQL template often aligns with multiple question templates. Next, we fill slots with the new database schema to generate numerous template (question, SQL) pairs. However, these templated questions can seem artificial since they directly insert table and column names. To make these questions more natural, we leverage GPT-3.5 to rephrase them based on $f$ manually crafted refinement examples. As showned in Figure~\ref{fig:augmentation}(b), each example comprises a triplet: [TEMPLATED QUESTION$_{i}$], [TEMPLATED SQL$_{i}$], and [REFINED QUESTION$_{i}$]. The end result is a new pair, [NEW REFINED QUESTION] and [NEW TEMPLATED SQL], which align more closely with typical text-to-SQL datasets.

\section{Usage of \model}\label{sec:usage}
We use \smodel by fine-tuning and few-shot in-context learning.
Formally, given training set $D = \{d_{1}, d_{2}, ..., d_{h}\}$ with $h$ text-to-SQL samples, each sample is a triplet $(d_{i}^{db}, d_{i}^{q}, d_{i}^{sql})$ denoting the database, question, and associated SQL query. For a test sample $t$, consisting of database $t^{db}$ and question $t^{q}$, we aim to generate SQL query $t^{sql}$. 
Before employing \model, we first convert each database in the text-to-SQL sample into its corresponding database prompt (see Section~\ref{sec:db_prompt}). Thus, each training sample is represented as a triplet: $(d_{i}^{dbp}, d_{i}^{q}, d_{i}^{sql})$, with $d_{i}^{dbp}$ being the prompt for database $d_{i}^{db}$. Similarly, each test sample is transformed into the pair $(t^{dbp}, t^{q})$, where $t^{dbp}$ serves as the prompt for database $t^{db}$.

\subsection{Supervised Fine-Tuning}
Given a substantial collection of training data, supervised fine-tuning (SFT) is a preferred option as it allows rapid adaptation to the specific data distribution. First, we form the input sequence as the combination of the database prompt and the question. Then, \smodel is optimized to predict the desired SQL query based on this input sequence. Therefore, the learning objective of SFT \smodel is:
\begin{equation}
    Loss = \frac{1}{h} \sum_{i=1}^{h}p(d_{i}^{sql}|d_{i}^{dbp}, d_{i}^{q}).
\end{equation}
After the fine-tuning process, for the given test sample, the refined \smodel can readily produce the SQL query using the combined inputs of $t^{dbp}$ and $t^{q}$.

\subsection{Few-shot In-Context Learning}
In cases where fine-tuning is impractical, we can utilize the model's built-in text-to-SQL capabilities without any fine-tuning. The efficacy of few-shot learning isn't solely based on the model's intrinsic capabilities; it's also influenced by the examples used~\cite{wei2022@cot, liu2022@what, zhang2022@active}. Thus we employ an efficient retriever to source valuable demonstrations.

Our goal is to select $K$ useful examples from the dataset $D$ to aid the model in predicting the right SQL query. A basic way is to check the semantic relevance between the test question, $t^{q}$, and all training questions, $\{d_{1}^{q}, d_{2}^{q}, ..., d_{h}^{q}\}$, evaluating the top-K training samples that match best. However, this can overly prioritize entities, leading to demonstrations that simply reflect the test question's entities. For instance, a question asking for singers born in 1948 or 1949 might fetch a training question about an artist who sang the most songs, due to shared references to ``\texttt{singers and songs}''.

To avoid overemphasizing entities, we focus on the question's core structure by stripping its entities. For example, we aim to enable the most suitable demonstration question for the question about singers born in 1948 or 1949 as ``\texttt{Show the names of members from either 'United States' or 'Canada'}'' without being limited to ``\texttt{singers and songs}''.

Formally, we define the similarity score between the test question $t^{q}$ and the training question $d_{i}^{q}$ as:
\begin{equation}
    \label{eq:sim}
    max(sentsim(t^{q}, d_{i}^{q}), sentsim(t^{qs}, d_{i}^{qs})),
\end{equation}

\noindent where $t^{qs}$ and $d_{i}^{qs}$ represent the extracted question patterns from $t^{q}$ and $d_{i}^{q}$, respectively. Using nltk's tool, we identify and remove entities from questions to get their patterns. We then use SimCSE~\cite{gao2021@simcse}, a sentence similarity model, to compute sequence similarities. We term this enhanced retrieval approach the ``question-pattern-aware demonstration retriever''.

Finally, after selecting the $K$ most relevant examples, we merge them with the test sample, creating a unified input sequence, which is then fed into the pre-trained model to derive the SQL query.
\section{Experiments}\label{sec:exp}
\subsection{Experimental Setup}
\subsubsection{Datasets.}\label{sec:datasets} We perform main experiments on two English text-to-SQL benchmarks: Spider~\cite{yu2018@spider} and BIRD~\cite{li2023@bird}. 
We also assess our models' robustness across four more challenging benchmarks: Spider-DK~\cite{gan2021@spider-dk}, Spider-Syn~\cite{gan2021@spider-syn}, Spider-Realistic~\cite{deng2021@spider-realistic}, and Dr.Spider~\cite{chang2023@drspider}. 
Following experimental settings in previous studies~\cite{gu2023@zeronl2sql, li2023@resdsql}, we utilize Spider as the training set and evaluate our models against these robustness-diagnostic test sets. 
Moreover, we manually created two databases from financial and academic domains, named Bank-Financials and Aminer-Simplified.

\stitle{Spider} offers a training set comprising 8,659 samples, a development set with 1,034 samples, and a hidden test set. The training portion encompasses 7,000 manually annotated samples, supplemented by an additional 1,659 samples sourced from six previous text-to-SQL datasets: Restaurants~\cite{tang2001@restaurants-1, popescu2003@restaurants-2}, GeoQuery~\cite{zelle1996@geoquery}, Scholar~\cite{iyer2017@scholar}, Academic~\cite{li2014@academic}, IMDB, and Yelp~\cite{yaghmazadeh2017@imdb}. Spider contains 200 databases that cover 138 diverse domains. However, due to the hardware constraints of the Spider submission platform\footnote{The Spider benchmark utilizes CodaLab Worksheets as its submission platform, which can be accessed at \url{https://worksheets.codalab.org/home}. This platform offers GPUs equipped with 12GB of memory. However, our best model, SFT \model-15B, requires at least 35GB of GPU memory to perform inference.}, we can not evaluate our models against its test set. Consequently, for Spider, primary evaluations utilize its publicly available development set.

\stitle{BIRD} comprises a training set of 9,428 samples, a development set with 1,534 samples, and a hidden test set. BIRD encompasses 95 databases, cumulatively accounting for 33.4GB across 37 distinct professional domains. BIRD is more challenging, with each of BIRD's databases containing around 549K rows on average, in contrast to Spider's limited capacity of just 2,000 rows. Moreover, BIRD offers \textbf{external knowledge (EK)} for specific samples to facilitate the generation of the right SQL query. Since it's impractical for users to supply such external knowledge, our evaluations on BIRD are categorized into two scenarios: with and without external knowledge. When the external knowledge is used, we simply integrate it with the question, yielding an enriched input prompt. We provide our code and models to the official organizers of BIRD for evaluation on their hidden test set.

\stitle{Spider-DK, Spider-Syn, Spider-Realistic} are variants derived from the original Spider dataset. They are specifically designed to mimic questions that could be posed by users in real-world scenarios. \textbf{Dr.Spider}, also a Spider derivative, incorporates 17 distinct perturbations across questions, databases, and SQL queries to holistically evaluate the robustness of text-to-SQL models. Specifically, Dr.Spider comprises 3 test sets with database perturbation, 9 test sets reflecting question perturbation, and 5 test sets with SQL perturbation.

\stitle{Bank-Financials} derives from a finance database containing 4 tables, with the largest table containing 65 columns (see Figure~\ref{fig:text2sql_example}). We generate a training set comprising 4,901 samples for the new finance database by using the data augmentation method proposed in Section~\ref{sec:adaption}. For evaluation, we further manually annotate 91 real-world questions as the test set. \textbf{Aminer-Simplified} originates from an academic database, sampled from a large-scale academic graph, AMiner~\cite{tang2008@arnetminer, zhang2023@oag2023, zhang2019@oag2019, sinha2015@mas}. We follow the same procedure as Bank-Financials and obtain a training set with 5,427 samples and a test set with 97 samples. Both training sets of these two datasets are derived from only 30 manually annotated samples.

\subsubsection{Evaluation Metrics.} (1) For Spider-family benchmarks (including Spider, Spider-DK, Spider-Syn, Spider-Realistic, and Dr-Spider), we consider two prevalent evaluation metrics: execution accuracy (EX)~\cite{yu2018@spider} and test-suite accuracy (TS)~\cite{zhong2020@test-suites}. The EX metric evaluates whether the predicted and ground-truth SQL queries yield the same execution results on the database. However, there can be instances where EX gives false positives — situations where an incorrect SQL prediction coincidentally produces the same output as the correct query~\cite{zhong2020@test-suites}. To counteract this, the test-suite accuracy was developed. It assesses if the generated SQL query consistently passes the EX evaluations across multiple database instances, which are derived from automated database augmentations. Due to its proficiency in reducing false positives, TS stands out as a more trustworthy metric than EX. It's worth noting that for Spider-DK and Dr.Spider, the TS script lacks test suites for their respective databases. As a result, we exclusively adopt the EX metric for them. (2) The BIRD benchmark primarily relies on execution accuracy (EX) as its evaluation metric, because the databases in BIRD typically contain a large number of values, minimizing the chances of false positives. Additionally, BIRD introduces the valid efficiency score (VES) to assess the execution efficiency of accurately generated SQL queries. Unlike EX, in VES, the score for an accurately generated SQL query is no longer 1 but is determined by the execution time of the ground truth SQL query divided by the execution time of the predicted SQL query. Thus, if the execution efficiencies are the same, the scores for EX and VES will be identical. However, if the predicted SQL query executes faster than the ground truth SQL query, the VES value will exceed that of EX. In practice, each correct predicted SQL query and its ground truth counterpart are executed 100 times, with their run times recorded. Yet, preliminary experiments indicated that VES could be highly susceptible to fluctuations based on varying hardware, software, and system status. Hence, for BIRD, EX serves as the stable and dependable metric.

\subsubsection{Evaluation Settings.} 
We evaluate \smodel under both few-shot, in-context learning and supervised fine-tuning scenarios. The few-shot in-context learning provides insight into the inherent SQL generation capabilities of language models, as it doesn't involve any fine-tuning. Instead, the models rely on their pre-training knowledge to address the users' questions. Then, to assess the alignment ability of \smodel, we fully fine-tune it using the training set and evaluate the fine-tuned version on different dev and test sets.

\subsubsection{Implementation details.}\label{sec:impl} In our experiments, we utilize SQLite to host and manage all databases. The training and evaluation processes for the schema item classifier are conducted in accordance with~\cite{li2023@resdsql}. Specifically, for each dataset, we train a classifier on its respective training set and assess its classification accuracy on the development set, utilizing the AUC metric for evaluation. The results of this evaluation are presented in Table~\ref{tab:auc}. It is observed that the AUC scores for Spider consistently surpass those for BIRD (and BIRD w/ EK). We hypothesize that this disparity is stemmed from the prevalence of ambiguous schemas in BIRD. Additionally, incorporating external knowledge enhances classification accuracy in BIRD, especially when the external knowledge directly highlights tables and columns relevant to the question. 
In supervised \model fine-tuning, we set $top_{k1}$ and $top_{k2}$ to 6 and 10. For few-shot in-context learning, to accommodate more demonstrations in the input prompt, these are adjusted to 5 and 6. We then fine-tune \smodel for 4 epochs with a batch size of 128, a learning rate of $5e^{-6}$, and a max context length of 4,096. SFT model performance might be enhanced with hyperparameter tuning. The learning rate has a cosine decay and a linear warmup for the initial 5\% of training. Other optimization settings align with Section~\ref{sec:pt_details}. For generation, a beam search produces 4 SQL candidates, picking the first executable one as the outcome.

\begin{table}[!t]
    \centering
    \caption{Table and column AUC scores for the trained schema item classifiers.}
    \vspace{-1em}
    \begin{tabular}{lccc}
        \toprule
         & Spider & BIRD & BIRD w/ EK \\
        \midrule
        Table AUC & \textbf{0.991} & 0.966 & 0.976 \\
        Column AUC & \textbf{0.993} & 0.943 & 0.957 \\
        \bottomrule
    \end{tabular}
    \label{tab:auc}
    \vspace{-1em}
\end{table}

\begin{table*}[!t]
    \centering
    \small
    \caption{In-context learning performance on  Spider's and BIRD's dev sets using 1-shot, 3-shot, and 5-shot settings. Due to space constraints, we only present the TS (\%) results for Spider and the EX (\%) results for BIRD and BIRD w/ EK (with external knowledge). The top performance is emphasized in bold, while the runner-up is underlined.}
    \vspace{-1em}
    \begin{tabular}{l|ccc|ccc|ccc}
        \toprule
        LLMs & \multicolumn{3}{c|}{Spider's dev (TS\%)} & \multicolumn{3}{c|}{BIRD's dev (EX\%)} & \multicolumn{3}{c}{BIRD's dev w/ EK (EX\%)} \\
         & 1-shot & 3-shot & 5-shot & 1-shot & 3-shot & 5-shot & 1-shot & 3-shot & 5-shot \\
        \midrule
        StarCoderBase-1B~\cite{li2023@starcoder} & 43.7 & 46.8 & 48.6 & 17.21 & 18.84 & 20.08 & 15.78 & 21.90 & 22.69 \\
        StarCoderBase-3B~\cite{li2023@starcoder} & 58.5 & 60.0 & 60.8 & 23.01 & 26.01 & 26.27 & 27.18 & 32.72 & 36.31 \\
        CodeGen-mono-6B~\cite{nijkamp2023@codegen} & 44.6 & 46.7 & 45.0 & 14.47 & 16.69 & 15.91 & 13.10 & 15.78 & 16.04 \\
        StarCoderBase-7B~\cite{li2023@starcoder} & 59.7 & 63.1 & 64.6 & 26.66 & 29.79 & 30.44 & 32.99 & 39.24 & 40.61 \\
        CodeGen2-7B~\cite{nijkamp2023@codegen2} & 46.8 & 49.8 & 50.8 & 18.38 & 18.90 & 19.30 & 16.56 & 19.88 & 19.56 \\
        Llama2-7B~\cite{touvron2023@llama2} & 34.9 & 39.3 & 40.2 & 12.45 & 16.30 & 15.12 & 15.25 & 19.04 & 19.88 \\
        Llama2-13B~\cite{touvron2023@llama2} & 45.4 & 48.5 & 47.6 & 16.88 & 20.34 & 20.47 & 21.06 & 25.81 & 25.36 \\
        StarCoderBase-15B~\cite{li2023@starcoder} & 63.5 & 67.7 & 70.0 & 29.86 & 31.55 & 33.77 & 35.40 & 39.24 & 41.20 \\
        StarCoder-15B~\cite{li2023@starcoder} & 63.8 & 67.6 & 69.6 & 29.86 & 32.99 & 33.64 & 35.46 & 40.09 & 42.44 \\
        StarCoderPlus-15B~\cite{li2023@starcoder} & 58.3 & 63.3 & 65.5 & 27.51 & 30.57 & 31.68 & 32.86 & 39.31 & 41.53 \\
        CodeGen-mono-16B~\cite{nijkamp2023@codegen} & 50.9 & 52.8 & 52.4 & 20.34 & 21.58 & 22.23 & 22.95 & 25.42 & 23.92 \\
        CodeGen2-16B~\cite{nijkamp2023@codegen2} & 51.6 & 55.9 & 57.4 & 21.45 & 22.69 & 23.14 & 23.01 & 25.42 & 25.23 \\
        \midrule
        \model-1B & 55.7 \scriptsize ($\uparrow$12.0) & 57.4 \scriptsize ($\uparrow$10.6) & 59.1 \scriptsize ($\uparrow$10.5) & 22.23 \scriptsize ($\uparrow$5.02) & 25.42 \scriptsize ($\uparrow$6.58) & 27.18 \scriptsize ($\uparrow$7.1) & 25.62 \scriptsize ($\uparrow$9.84) & 29.47 \scriptsize ($\uparrow$7.57) & 31.03 \scriptsize ($\uparrow$8.34) \\
        \model-3B & 64.6 \scriptsize ($\uparrow$6.1) & 67.8 \scriptsize ($\uparrow$7.8) & 69.7 \scriptsize ($\uparrow$8.9) & 29.07 \scriptsize ($\uparrow$6.06) & 31.23 \scriptsize ($\uparrow$5.22) & 31.81 \scriptsize ($\uparrow$5.54) & 35.59 \scriptsize ($\uparrow$8.41) & 39.18 \scriptsize ($\uparrow$6.46) & 41.85 \scriptsize ($\uparrow$5.54) \\
        \model-7B & \underline{66.0} \scriptsize ($\uparrow$6.3) & \underline{69.8} \scriptsize ($\uparrow$6.7) & \underline{71.8} \scriptsize ($\uparrow$7.2) & \underline{30.77} \scriptsize ($\uparrow$4.11) & \underline{33.44} \scriptsize ($\uparrow$3.65) & \underline{34.49} \scriptsize ($\uparrow$4.05) & \underline{37.29} \scriptsize ($\uparrow$4.30) & \underline{41.98} \scriptsize ($\uparrow$2.74) & \underline{44.26} \scriptsize ($\uparrow$3.65) \\
        \model-15B & \textbf{69.3} \scriptsize ($\uparrow$5.5) & \textbf{71.5} \scriptsize ($\uparrow$3.9) & \textbf{73.4} \scriptsize ($\uparrow$3.8) & \textbf{34.09} \scriptsize ($\uparrow$4.23) & \textbf{35.14} \scriptsize ($\uparrow$2.15) & \textbf{37.03} \scriptsize ($\uparrow$3.39) & \textbf{39.57} \scriptsize ($\uparrow$4.11) & \textbf{43.48} \scriptsize ($\uparrow$3.39) & \textbf{45.44} \scriptsize ($\uparrow$3.00) \\
        \bottomrule
    \end{tabular}
    \label{tab:few_shot}
\end{table*}

\subsubsection{Enviroments.}\label{sec:env} Our experiments are conducted using PyTorch 1.13.1 on a computer running the CentOS Linux 7 operating system, equipped with 8 NVIDIA A800 80GB GPUs, an Intel(R) Xeon(R) Platinum 8358 CPU, and 1024GB of RAM.

\subsubsection{Baselines.} For few-shot in-context learning, our objective is to gauge the inherent SQL generation capabilities of \model. Therefore, our baselines consist of widely recognized open-source language models, such as those from the StarCoder~\cite{li2023@starcoder}, CodeGen~\cite{nijkamp2023@codegen, nijkamp2023@codegen2}, and Llama2~\cite{touvron2023@llama2}. For supervised fine-tuning, almost all baselines are drawn from the SOTA text-to-SQL approaches listed on the official leaderboards of benchmarks. We additionally fine-tune Llama2-7B and 13B as our competitive baselines.

\subsection{Evaluation on Few-shot In-Context Learning}
Table~\ref{tab:few_shot} shows the results of few-shot in-context learning evaluations on the Spider and BIRD benchmarks. We increase the demonstrations from 1 to 3 to 5. For a fair comparison, all models use our few-shot in-context learning framework with consistent input prompts. 
When comparing various versions of StarCoder (\textit{i.e.}, before incremental pre-training) with our \smodel (\textit{i.e.}, after incremental pre-training), it's clear that incremental pre-training greatly improves StarCoder's SQL generation capability.
\textit{Consequently, \smodel-15B emerges as the leading open-source pre-trained language model in SQL generation.} Furthermore, smaller models exhibit a more pronounced improvement compared to larger models. This observation underscores our initial hypothesis that smaller models, due to their constrained parameters, may not be optimally trained for SQL-related tasks.

\begin{table}[!t]
    \centering
    \small
    \caption{Evaluation of SFT \smodel on Spider's dev set.}
    \vspace{-1em}
    \begin{tabular}{lcc}
        \toprule
        Methods & EX\% & TS\% \\
        \midrule
        \multicolumn{3}{c}{Fine-tuning-based methods} \\
        \midrule
        T5-3B + PICARD~\cite{scholak2021@picard} & 79.3 & 69.4 \\
        RESDSQL-3B + NatSQL~\cite{li2023@resdsql} & 84.1 & 73.5 \\
        Graphix-T5-3B + PICARD~\cite{li2023@graphix} & 81.0 & 75.0 \\
        Fine-tuned SQL-PaLM~\cite{sun2023@sqlpalm} & 82.8 & 78.2 \\
        SFT Llama2-7B~\cite{touvron2023@llama2} & 77.8 & 73.0 \\
        SFT Llama2-13B~\cite{touvron2023@llama2} & 81.6 & 76.6 \\
        \midrule
        \multicolumn{3}{c}{Prompting-based methods} \\
        \midrule
        GPT-4 (few-shot)~\cite{pourreza2023@dinsql} & 76.8 &  67.4 \\
        C3 + ChatGPT~\cite{dong2023@c3} & 81.8 & 71.4 \\
        Self-Debug + Codex davinci~\cite{chen2023@selfdebug} & 84.1 & - \\
        DIN-SQL + GPT-4~\cite{pourreza2023@dinsql} & 82.8 & 74.2 \\
        DAIL-SQL + GPT-4~\cite{gao2023@dailsql} & 83.1 & 76.6 \\
        SQL-PaLM (few-shot)~\cite{sun2023@sqlpalm} & 82.7 & 77.3 \\
        DAIL-SQL + GPT-4 + Self-Consistency~\cite{gao2023@dailsql} & 83.6 & 76.2 \\
        \midrule
        \multicolumn{3}{c}{Ours} \\
        \midrule
        SFT \model-1B & 77.9 & 72.2 \\
        SFT \model-3B & 83.4 & 78.1 \\
        SFT \model-7B & \textbf{85.4} & \textbf{80.3} \\
        SFT \model-15B & \underline{84.9} & \underline{79.4} \\
        \bottomrule
    \end{tabular}
    \label{tab:fine_tune_spider}
    \vspace{-1em}
\end{table}

\subsection{Evaluation on Supervised Fine-Tuning}\label{sec:eval_sft}
Table~\ref{tab:fine_tune_spider} presents the evaluation results in EX and TS on Spider's dev set. Remarkably, SFT \model-3B outperforms the leading GPT-4-based method (\emph{i.e.}, DIN-SQL and DAIL-SQL), illustrating the potential of smaller models to excel as text-to-SQL parsers after fine-tuning. Furthermore, SFT \model-7B and SFT \model-15B achieve new SOTA performance on Spider's development set. However, SFT \model-7B exhibits a marginal advantage over SFT \model-15B, suggesting potential overfitting of \model-15B to the Spider training data, which might slightly compromise its generalization to the development set. This phenomenon indicates that a larger model doesn't always guarantee better fine-tuning results.

Table~\ref{tab:fine_tune_bird} presents the evaluation results in EX and VES on  BIRD's development and hidden test sets. Given BIRD's higher complexity compared to Spider, our approach yields more significant improvements over the baseline methods. Without using external knowledge, SFT \model-15B significantly outperforms ChatGPT + COT, improving from 28.95\% to 52.15\% in EX on the test set, a notable increase of 23.20\%. When incorporating external knowledge (w/ EK), both SFT \model-7B and SFT \model-15B show clear EX improvements of 3.35\% and 4.47\%, respectively, on the test set compared to the previous SOTA text-to-SQL method, DIN-SQL + GPT-4. However, even though SFT \model-15B outperforms SFT \model-7B, the margin between them remains minimal, especially considering the doubled parameter size in the former. This suggests that \model-7B offers an optimal trade-off between computational efficiency and text-to-SQL capabilities. Additionally, the VES metric surpassing EX signifies that our models are more capable than human experts in producing execution-efficient SQL queries. At the time of writing, SFT \model-15B and SFT \model-7B hold the top positions on BIRD's official leaderboard. 

\begin{table}[!t]
    \centering
    \caption{Evaluation of SFT \smodel on BIRD's dev/test sets.}
    \vspace{-1em}
    \scalebox{0.73}{
    \begin{tabular}{@{}l|cc|cc|cc|cc@{}}
        \toprule
         & \multicolumn{2}{c|}{\textbf{Dev}} & \multicolumn{2}{c|}{\textbf{Dev w/ EK}} & \multicolumn{2}{c|}{\textbf{Test}} & \multicolumn{2}{c}{\textbf{Test w/ EK}} \\
        \midrule
        Methods & EX\% & VES\% & EX\% & VES\% & EX\% & VES\% & EX\% & VES\% \\
        \midrule
        \multicolumn{9}{c}{Baseline methods} \\
        \midrule
        Fine-tuned T5-3B~\cite{li2023@bird} & 10.37 & 13.62 & 23.34 & 25.57 & 11.17 & 15.17 & 24.05 & 27.80 \\
        PaLM-2~\cite{li2023@bird} & - & - & 27.38 & - & - & - & 33.04 & - \\
        Codex 175B~\cite{li2023@bird} & 25.42 & 33.37 & 34.35 & 43.41 & 24.86 & 35.40 & 36.47 & 41.60 \\
        ChatGPT~\cite{li2023@bird} & 24.05 & 27.97 & 42.24 & - & 26.77 & 36.68 & 39.30 & 51.40 \\
        ChatGPT + CoT~\cite{li2023@bird} & 25.88 & 32.33 & 36.64 & 42.30 & 28.95 & 49.69 & 40.08 & 56.56 \\
        Claude-2~\cite{li2023@bird} & - & - & 42.70 & - & - & - & 49.02 & - \\
        GPT-4~\cite{li2023@bird} & - & - & 49.15 & - & - & - & 54.89 & 60.77 \\
        DIN-SQL + GPT-4~\cite{pourreza2023@dinsql} & - & - & 50.72 & 58.79 & - & - & 55.90 & 59.44 \\
        DAIL-SQL + GPT-4~\cite{gao2023@dailsql} & - & - & 54.76 & 56.08 & - & - & 57.41 & 61.95 \\
        SFT Llama2-7B~\cite{touvron2023@llama2} & 35.53 & 36.09 & 45.37 & 46.98 & - & - & - & - \\
        SFT Llama2-13B~\cite{touvron2023@llama2} & 41.85 & 44.00 & 53.91 & 58.77 & - & - & - & - \\
        \midrule
        \multicolumn{9}{c}{Ours} \\
        \midrule
        SFT \model-1B & 38.46 & 41.77 & 50.46 & 51.07 & - & - & - & - \\
        SFT \model-3B & 43.42 & 44.55 & 55.02 & 56.54 & - & - & - & - \\
        SFT \model-7B & \underline{45.24} & \underline{48.13} & \underline{57.17} & \underline{58.80} & \underline{50.25} & \underline{54.84} & \underline{59.25} & \underline{63.62} \\
        SFT \model-15B & \textbf{47.91} & \textbf{49.60} & \textbf{58.47} & \textbf{59.87} & \textbf{52.15} & \textbf{56.99} & \textbf{60.37} & \textbf{64.22} \\
        \bottomrule
    \end{tabular}
    }
    \label{tab:fine_tune_bird}
    \vspace{-1em}
\end{table}

\begin{table}[!t]
    \centering
    \caption{Evaluation of SFT \smodel on Spider variants.}
    \vspace{-1em}
    \scalebox{0.85}{
    \begin{tabular}{@{}lccccc@{}}
        \toprule
         & \multicolumn{2}{c}{\textbf{Spider-Syn}} & \multicolumn{2}{c}{\textbf{Spider-Realistic}} & \textbf{Spider-DK}\\
        \midrule
        Methods & EX\% & TS\% & EX\% & TS\% & EX\% \\
        \midrule
        T5-3B + PICARD~\cite{scholak2021@picard} & 69.8 & 61.8 & 71.4 & 61.7 & 62.5 \\
        RESDSQL-3B + NatSQL~\cite{li2023@resdsql} & \underline{76.9} & 66.8 & 81.9 & 70.1 & 66.0 \\
        ChatGPT~\cite{liu2023@chatgptzeroshot} & 58.6 & 48.5 & 63.4 & 49.2 & 62.6 \\
        SQL-PaLM (Few-shot)~\cite{sun2023@sqlpalm} & 74.6 & 67.4 & 77.6 & 72.4 & 66.5 \\
        SQL-PaLM (Fine-tuned)~\cite{sun2023@sqlpalm} & 70.9 & 66.4 & 77.4 & 73.2 & 67.5 \\
        \midrule
        SFT \model-1B & 66.5 & 59.3 & 70.9 & 61.8 & 64.7 \\
        SFT \model-3B & 75.7 & 69.0 & 79.9 & 74.4 & \underline{71.8} \\
        SFT \model-7B & \underline{76.9} & \textbf{70.0} & \underline{82.9} & \textbf{77.2} & \textbf{72.0} \\
        SFT \model-15B & \textbf{77.0} & \underline{69.4} & \textbf{83.1} & \underline{75.6} & 70.7 \\
        \bottomrule
    \end{tabular}
    }
    \label{tab:spider_robustness}
\end{table}

\begin{table*}[!t]
    \centering
    \small
    \caption{Evaluation of SFT \smodel on Dr.Spider in terms of EX (\%). ``DB'', ``NLQ'', and `SQL'' denote perturbations in the database, user's questions, and SQL side respectively. Macro-average is computed across various perturbations.}
    \vspace{-1em}
    \begin{tabular}{llcccccccc}
        \toprule
        \textbf{Type} & \textbf{Perturbation} & \footnotesize \textbf{\#Samples} & \footnotesize \textbf{SMBOP}~\cite{rubin2020@smbop} & \footnotesize \textbf{RESDSQL-3B} & \footnotesize \textbf{ChatGPT} & \footnotesize \textbf{SFT} & \footnotesize \textbf{SFT} & \footnotesize \textbf{SFT} & \footnotesize \textbf{SFT} \\
        & & & & \footnotesize \textbf{+NatSQL}~\cite{li2023@resdsql} & \footnotesize \textbf{+ZeroNL2SQL}~\cite{gu2023@zeronl2sql} & \footnotesize \textbf{\model-1B}& \footnotesize \textbf{\model-3B} & \footnotesize \textbf{\model-7B} & \footnotesize \textbf{\model-15B} \\
        \midrule
        \multirow{4}{*}{DB} & schema-synonym & 2,619 & 53.9 & \underline{68.3} & \textbf{69.8} & 58.5 & 64.3 & 67.2 & 66.9 \\
         & schema-abbreviation & 2,853 & 59.0 & 70.0 & 74.8 & 68.6 & 75.0 & \underline{76.8} & \textbf{78.7} \\
         & DBcontent-equivalence & 382 & 37.2 & 40.1 & \textbf{56.8} & \underline{53.9} & 47.9 & 46.9 & 47.6 \\
         & Average & - & 50.0 & 59.4 & \textbf{67.1} & 60.3 & 62.4 & 63.6 & \underline{64.4} \\
        \midrule
        \multirow{10}{*}{NLQ} & keyword-synonym & 953 & 64.3 & 72.4 & \textbf{74.0} & 60.9 & 70.9 & 73.0 & \underline{73.5} \\
         & keyword-carrier & 399 & 79.2 & 83.5 & 88.2 & 86.5 & 91.2 & \underline{91.5} & \textbf{91.7} \\
         & column-synonym & 563 & 48.7 & 63.1 & 62.7 & 56.0 & 60.0 & \underline{63.2} & \textbf{64.7} \\
         & column-carrier & 579 & 64.6 & 63.9 & 71.7 & 67.4 & 74.4 & \textbf{80.7} & \underline{79.1} \\
         & column-attribute & 119 & 58.0 & \textbf{71.4} & \underline{70.6} & 47.9 & 67.2 & 63.0 & 68.9 \\
         & column-value & 304 & 58.9 & \textbf{76.6} & 76.0 & 72.4 & 75.0 & 73.7 & \underline{76.3} \\
         & value-synonym & 506 & 29.1 & 53.2 & 70.6 & 59.7 & 67.0 & \textbf{72.7} & \underline{71.9} \\
         & multitype & 1,351 & 46.1 & 60.7 & 66.4 & 57.5 & 66.5 & \textbf{69.5} & \underline{69.4} \\
         & others & 2,819 & 73.7 & 79.0 & 79.4 & 74.9 & 78.5 & \textbf{81.5} & \underline{81.2} \\
         & Average & - & 58.1 & 69.3 & 73.2 & 64.8 & 72.3 & \underline{74.3} & \textbf{75.2} \\
        \midrule
        \multirow{6}{*}{SQL} & comparison & 178 & 65.2 & \textbf{82.0} & 73.6 & 60.7 & 69.7 & \underline{77.5} & 71.9 \\
         & sort-order & 192 & 76.6 & \textbf{85.4} & 80.2 & 69.8 & 79.2 & 81.8 & \underline{84.9} \\
         & nonDB-number & 131 & 71.8 & 85.5 & \textbf{92.4} & 84.7 & 87.8 & \underline{90.1} & 84.0 \\
         & DB-text & 911 & 63.1 & 74.3 & \textbf{80.7} & 67.1 & 77.2 & \underline{80.5} & \textbf{80.7} \\
         & DB-number & 410 & 84.4 & \textbf{88.8} & \underline{86.1} & 80.5 & 85.1 & 84.9 & 85.9 \\
         & Average & - & 72.2 & \textbf{83.2} & 82.6 & 72.6 & 79.8 & \underline{83.0} & 81.5 \\
        \midrule
        All & Global average & - & 60.8 & 71.7 & 74.9 & 66.3 & 72.8 & \underline{75.0} & \textbf{75.1} \\
        \bottomrule
    \end{tabular}
    \label{tab:drspider}
\end{table*}

\subsection{Evaluation on Robustness Benchmarks}
Table~\ref{tab:spider_robustness} evaluates SFT \smodel on three Spider variants for robustness: Spider-DK, Spider-Syn, and Spider-Realistic. Notably, SFT \model-7B exhibits exceptional performance, achieving gains of 2.6\% on Spider-Syn (67.4\% to 70.0\%), 4.0\% on Spider-Realistic (73.2\% to 77.2\%), and 4.5\% on Spider-DK (67.5\% to 72.0\%), comparing with the best baselines. 
Even the SFT \model-3B outperforms previous SOTA methods across all the datasets. Considering that SFT \smodel is trained on Spider but tested on its variants, these results highlight the model's impressive generalization capability in challenging distribution shift scenarios.

For a more comprehensive evaluation of the robustness of SFT \model, we further test SFT \smodel on Dr.Spider. The results can be found in Table~\ref{tab:drspider}. First, for the database (DB) perturbation, \smodel slightly lags behind ChatGPT + ZeroNL2SQL, mainly due to the \textit{DBcontent-equivalence} perturbation. ChatGPT + ZeroNL2SQL uses a dense retriever for better semantic accuracy, but it's resource-intensive, needing more disk space and computation time. To maintain efficiency and real-world applicability, we opt for a sparse retriever. In the natural language question perturbation, both SFT \model-7B and SFT \model-15B outperform the previous best, ChatGPT + ZeroNL2SQL, scoring 74.3\% and 75.2\% versus 73.2\%. This suggests our models have a better grasp of question semantics, leading to more accurate SQL queries. For SQL perturbations, our models slightly lag behind RESDSQL-3B + NatSQL. The latter's intermediary SQL representation, NatSQL, is more simple than SQL and aids its performance. However, its syntax is limited to Spider, making it less adaptable. In global average performance, SFT \model-7B and SFT \model-15B slightly surpass the prior best, ChatGPT + ZeroNL2SQL, which is tailored for robust text-to-SQL. Overall, even without a specific robustness design, our models frequently excel against methods built especially for text-to-SQL resilience. 

\subsection{Ablation studies}
We conduct an extensive ablation study on Spider and BIRD under the 3-shot in-context learning setting to evaluate the impact of each key component. The results are shown in Table~\ref{tab:ablations}.

\stitle{Demonstration Retriever.} 
Using only question similarity for retrieving demonstrations (-w/o pattern similarity) results in a performance drop on Spider but less so on BIRD. This could be because Spider has less question variety than BIRD, making it easier to group similar text-to-SQL samples based on their patterns. Then, when replacing our demonstration retrieval strategy with a purely random selection (-w/o demonstration retriever), performance decreases in most scenarios.

\stitle{Schema Filter and Value Retriever.} 
We explore the effects of the schema filter and value retriever on performance. Without the schema filter, there's both a drop in performance and a slower generation speed due to longer input sequences. Omitting the value retriever also leads to a marked decrease in text-to-SQL performance, especially on the BIRD benchmark, highlighting its crucial role in generating SQL query predicates.

\stitle{Metadata.} 
We perform ablations on metadata. As per Table~\ref{tab:ablations}, column data types have a minor performance impact, possibly because the model infers types from column names and comments. Removing comments notably affects performance on the BIRD benchmark due to its ambiguous schemas. Additionally, both representative database values and primary/foreign keys are essential for performance on Spider and BIRD. The first offers insight into value format, and the latter helps in accurate \texttt{JOIN ON} clause generation.

\begin{table*}[!t]
    \centering
    \small
    \caption{Ablation studies on Spider's and BIRD's dev sets in the 3-shot in-context learning manner.}
    \vspace{-1em}
    \begin{tabular}{l|cccc|cccc}
        \toprule
         & \multicolumn{4}{c|}{\textbf{Spider's dev (TS\%)}} & \multicolumn{4}{c}{\textbf{BIRD's dev (EX\%)}} \\
         & \model-1B & \model-3B & \model-7B & \model-15B & \model-1B & \model-3B & \model-7B & \model-15B \\
        \midrule
         Original & 57.4 & 67.8 & 69.8 & 71.5 & 25.42 & 31.23 & 33.44 & 35.14 \\
        \midrule
        \multicolumn{9}{c}{Ablations on demonstration retriever} \\
        \midrule
         -w/o pattern similarity & 55.8\scriptsize ($\downarrow$-1.6) & 66.2\scriptsize ($\downarrow$-1.6) & 67.6\scriptsize ($\downarrow$-2.2) & 69.7\scriptsize ($\downarrow$-1.8) & 25.62\scriptsize ($\uparrow$+0.20) & 31.16\scriptsize ($\downarrow$-0.07) & 34.09\scriptsize ($\uparrow$+0.65) & 35.79\scriptsize ($\uparrow$+0.65) \\
         -w/o demonstration retriever & 56.1\scriptsize ($\downarrow$-1.3) & 66.8\scriptsize ($\downarrow$-1.0) & 69.6\scriptsize ($\downarrow$-0.2) & 71.4\scriptsize ($\downarrow$-0.1) & 24.25\scriptsize ($\downarrow$-1.17) & 30.18\scriptsize ($\downarrow$-1.05) & 32.86\scriptsize ($\downarrow$-0.58) & 35.53\scriptsize ($\uparrow$+0.39) \\
        \midrule
        \multicolumn{9}{c}{Ablations on schema filter and value retriever} \\
        \midrule
         -w/o schema filter & 55.0\scriptsize ($\downarrow$-2.4) & 65.4\scriptsize ($\downarrow$-2.4) & 69.0\scriptsize ($\downarrow$-0.8) & 70.2\scriptsize ($\downarrow$-1.3) & 23.53\scriptsize ($\downarrow$-1.89) & 30.64\scriptsize ($\downarrow$-0.59) & 33.83\scriptsize ($\uparrow$+0.39) & 33.57\scriptsize ($\downarrow$-1.57) \\
         -w/o value retriever & 57.2\scriptsize ($\downarrow$-0.2) & 66.7\scriptsize ($\downarrow$-1.1) & 69.6\scriptsize ($\downarrow$-0.2) & 71.1\scriptsize ($\downarrow$-0.4) & 22.23\scriptsize ($\downarrow$-3.19) & 29.27\scriptsize ($\downarrow$-1.96) & 30.96\scriptsize ($\downarrow$-2.48) & 33.31\scriptsize ($\downarrow$-1.83) \\
        \midrule
        \multicolumn{9}{c}{Ablations on metadata} \\
        \midrule
         -w/o column data types & 56.3\scriptsize ($\downarrow$-1.1) & 66.9\scriptsize ($\downarrow$-0.9) & 69.4\scriptsize ($\downarrow$-0.4) & 71.1\scriptsize ($\downarrow$-0.4) & 24.71\scriptsize ($\downarrow$-0.71) & 30.83\scriptsize ($\downarrow$-0.4) & 33.83\scriptsize ($\uparrow$+0.39) & 35.66\scriptsize ($\uparrow$+0.52) \\
         -w/o comments & 57.7\scriptsize ($\uparrow$+0.3) & 67.0\scriptsize ($\downarrow$-0.8) & 69.2\scriptsize ($\downarrow$-0.6) & 71.0\scriptsize ($\downarrow$-0.5) & 24.71\scriptsize ($\downarrow$-0.71) & 29.92\scriptsize ($\downarrow$-1.31) & 32.33\scriptsize ($\downarrow$-1.11) & 34.03\scriptsize ($\downarrow$-1.11) \\
         -w/o representative values & 57.6\scriptsize ($\uparrow$+0.2) & 66.4\scriptsize ($\downarrow$-1.4) & 69.9\scriptsize ($\uparrow$+0.1) & 70.4\scriptsize ($\downarrow$-1.1) & 23.92\scriptsize ($\downarrow$-1.50) & 29.40\scriptsize ($\downarrow$-1.83) & 30.77\scriptsize ($\downarrow$-2.67) & 31.94\scriptsize ($\downarrow$-3.20) \\
         -w/o primary and foreign keys & 57.6\scriptsize ($\uparrow$+0.2) & 66.2\scriptsize ($\downarrow$-1.6) & 69.0\scriptsize ($\downarrow$-0.8) & 70.0\scriptsize ($\downarrow$-1.5) & 23.27\scriptsize ($\downarrow$-2.15) & 28.29\scriptsize ($\downarrow$-2.94) & 29.92\scriptsize ($\downarrow$-3.52) & 32.14\scriptsize ($\downarrow$-3.00) \\
        \bottomrule
    \end{tabular}
    \label{tab:ablations}
\end{table*}

\begin{table}[!t]
    \centering
    \footnotesize
    \caption{Automatic and human evaluation results on Bank-Financials and Aminer-Simplified.}
    \vspace{-1em}
    \begin{tabular}{lcccc}
        \toprule
         & \multicolumn{2}{c}{\textbf{Bank-Financials}} & \multicolumn{2}{c}{\textbf{Aminer-Simplified}} \\
        \midrule
        Methods & EX\% & HE\% & EX\% & HE\% \\
        \midrule
        RESDSQL-3B + NatSQL & 6.6 & 26.4 & 17.5 & 24.7 \\
        3-shot GPT-3.5 & 52.7 & 72.5 & 50.5 & 63.9 \\
        3-shot GPT-3.5 + comments & 57.1 & \underline{84.6} & \underline{51.5} & 62.8 \\
        DIN-SQL + GPT-4 & 26.4 & 79.1 & 50.5 & \textbf{67.0} \\
        \midrule
        SFT \model-7B using Spider & 11.0 & 73.6 & 27.8 & 36.1 \\
        SFT \model-7B using BIRD w/ EK & 12.1 & 79.1 & 34.0 & 41.2 \\
        3-shot \model-7B & 61.5 & 78.0 & 43.3 & 51.5 \\
        SFT \model-7B using aug. data & \textbf{71.4} & \textbf{85.7} & \underline{51.5} & \underline{64.9} \\
        SFT \model-7B using merged data & \underline{65.9} & \underline{84.6} & \textbf{53.6} & \textbf{67.0} \\
        \bottomrule
    \end{tabular}
    \label{tab:new_domain}
\end{table}

\subsection{Evaluation on Real-World Scenarios}\label{sec:eval_new_domain}
We evaluate \smodel on two new domain datasets: Bank-Financials and Aminer-Simplified. The primary challenge of Bank-Financials lies in the large number of columns in the database and the presence of ambiguous schema names. Aminer-Simplified poses a challenge due to its complex and intricate table-join relationships. 

For real-world deployment, we use the \model-7B model due to its balance between performance and efficiency. We use the schema item classifier trained on BIRD during inference to filter schemas in new databases, as it scores based on question-schema semantics and is adaptable across domains. Our baselines come from the prompting-based GPT-3.5. We provide GPT-3.5 with three random text-to-SQL samples from new databases. The input structure is: ``[DDL] + [Comments (optional)] + 3 instances of [Question, SQL] + [Test question]''. Given the EX metric's strictness, we also employ human evaluation (HE) for SQL query accuracy. For example, consider the question, ``\texttt{What is the abstract of 'Optical geometries'?}''. The human annotated ground truth SQL query is ``\texttt{SELECT abstract FROM Paper WHERE title = 'Optical geometries';}''. However, if the generated SQL query additionally selects the ``\texttt{title}'' column from the table, EX would judge it incorrect. Yet, human experts would consider the predicted SQL as valid since it provides the essential information requested by the user.

Considering available labeled data and computational resources, we offer the following pathways for using \smodel:

(1) For new databases without annotations, we can directly use the checkpoints fine-tuned on Spider and BIRD benchmarks for inference. Table~\ref{tab:new_domain} shows that such ``SFT \model-7B using Spider'' and ``SFT \model-7B using BIRD w/ EK'' have certain domain transfer capability. It should be noted that the large gap between EX and HE is attributed to the different annotation habits between benchmarks and our new datasets.

(2) With limited annotated samples, if computational resources are scarce, \smodel's few-shot learning can quickly adapt to new databases without parameter tuning. In our tests, \model-7B, using just three context demonstrations, could generate SQL queries for new databases (refer to ``3-shot \model-7B'' in Table~\ref{tab:new_domain}). If resources permit, we can use the bi-directional data augmentation strategy proposed in Section~\ref{sec:adaption} to produce ample training pairs for two new databases. Table~\ref{tab:new_domain}'s ``SFT \model-7B using aug. data'' shows that using these augmented data for fine-tuning \model-7B can greatly boost accuracy.
However, fine-tuning a separate model for each database has substantial real-world overheads. We explored merging training data from Spider, BIRD, Bank-Financials, and Aminer-Simplified to train a unified text-to-SQL model. Results show that this approach not only prevents performance drops but also improves performance, especially on the Aminer-Simplified dataset, as seen in Table~\ref{tab:new_domain}'s ``SFT \model-7B using merged data''.

\subsection{Latency and Deployment Requirements}\label{sec:latency}
One of the key benefits of smaller models is their enhanced inference speed. To illustrate, the prior SOTA prompting-based method, DIN-SQL+GPT-4, reports approximately 60 seconds of inference time per sample on the Spider dataset. In contrast, our SFT \model demonstrates significantly improved efficiency. The inference times for the 1B, 3B, 7B, and 15B variants are only 0.6, 0.9, 1.1, and 1.5 seconds respectively on the same dataset. This highlights the superior speed of our models. In addition to its efficiency, \model is also suitable for real-world deployment. More specifically, when operating in float16 precision, the SFT \model variants (1B, 3B, 7B, 15B) require GPU memory capacities of 10GB, 13GB, 20GB, and 35GB, respectively. Therefore, we can deploy \model on local machines, bypassing the need for the expensive GPT-4 API.
\section{Conlusion}
In this study, we have taken significant strides toward enhancing the landscape of open-source text-to-SQL models. With the introduction of \model, developers now have access to a range of specialized pre-trained language models to develop their text-to-SQL applications. We also open-source our collected SQL-focused corpus to the research community, which could pave the way for future innovations in SQL generation using incremental pre-training. In addition, we propose a comprehensive database prompt construction strategy and a novel bi-directional data augmentation method. This ensures that the model remains versatile and can adapt seamlessly to various domains. Finally, we conduct extensive evaluations across various text-to-SQL benchmarks. Our findings showcase that \smodel is the new SOTA pre-trained language model in the SQL generation capability and our SFT \smodel models achieve new SOTA accuracy and robustness on a wide range of text-to-SQL benchmarks. We hope our efforts, combined with the open source of our code, models, and data, will catalyze further exploration, adoption, and innovation in the domain of text-to-SQL. Beyond this, we're optimistic that this work could offer invaluable perspectives for deploying language models across other specialized domains.



\bibliographystyle{ACM-Reference-Format}
\bibliography{sample-base}

\end{document}